\documentclass[10pt,twocolumn,letterpaper]{article}

\usepackage{iccv}
\usepackage{graphicx}
\usepackage{amsmath}
\usepackage{amssymb}
\usepackage{multirow}
\usepackage{tkz-fct} 
\usepackage{pgfplots} 
\usepackage{xcolor} 

\usepackage{tikz} 
\usetikzlibrary{arrows.meta}

\usepackage{enumitem}
\graphicspath{{figures/}{imgs/}}

\usepackage[pagebackref=true,breaklinks=true,colorlinks,bookmarks=false]{hyperref}

\iccvfinalcopy 

\def\iccvPaperID{5825} 

\newcommand{\C}{\mathcal{C}}
\ificcvfinal\fi
\begin{document}

\title{Intra-frame Object Tracking by Deblatting}

\author{Jan Kotera\\
UTIA, Prague, Czech Republic\\
{\tt\small kotera@utia.cas.cz}
\and
Denys Rozumnyi\\
CMP, Prague, Czech Republic\\
{\tt\small rozumden@cmp.felk.cvut.cz}
\and
Filip \v{S}roubek\\
UTIA, Prague, Czech Republic\\
{\tt\small sroubekf@utia.cas.cz}
\and
Ji\v{r}\'\i{} Matas\\
CMP, Prague, Czech Republic\\
{\tt\small matas@cmp.felk.cvut.cz}
}

\maketitle

\begin{abstract}
Objects moving at high speed along complex trajectories often appear in videos, especially videos of sports. Such objects elapse non-negligible distance during exposure time of a single frame and therefore their position in the frame is not well defined. They appear as semi-transparent streaks due to the motion blur and cannot be reliably tracked by standard trackers.

We propose a novel approach called Tracking by Deblatting based on the observation that motion blur is directly related to the intra-frame trajectory of an object.
Blur is estimated by solving two intertwined inverse problems, blind deblurring and image matting, which we call deblatting. The trajectory is then estimated by fitting a piecewise quadratic curve, which models physically justifiable trajectories. As a result, tracked objects are precisely localized with higher temporal resolution than by conventional trackers.

The proposed TbD tracker was evaluated on a newly created dataset of videos with ground truth obtained by a high-speed camera using a novel Trajectory-IoU metric that generalizes the traditional Intersection over Union and measures the accuracy of the intra-frame trajectory. The proposed method outperforms baseline both in recall and trajectory accuracy.
\end{abstract}

\section{Introduction}
\label{sec:Intro}
The field of visual object tracking has progressed significantly in recent years. The area encompasses a wide range of problems, including single object model-free short-term tracking where a single target is localized in a video sequence given a single training example~\cite{otb,vot2016,VOT_TPAMI,vot2018}, long-term tracking methods requiring redetection and learning\cite{kalal2012tracking,uav_benchmark_eccv2016,moudgil2017long,tao2017tracking}, multi-target multi-camera tracking \cite{RistaniSZCT16}, multi-view methods \cite{KroegerDG14} and methods targeting specific objects such as cars \cite{betke2000real}, humans \cite{mittal2003m}, or animals \cite{fry2000tracking}. Many variants of the problems have been considered~--~static or dynamic cameras or environments, RGBD input, use of inertial measurement units to name a few.

Recently, Rozumnyi~\etal~\cite{fmo} have shown that the performance of standard state-of-the-art trackers drops significantly when applied to Fast Moving Objects (FMO), apparently due to the effect of blur~--~such objects appear only as semi-transparent streaks. Examples of applications with FMOs include tracking of balls and ball-like objects in sports videos, particles in scientific experiments, and flying birds and insects.

\begin{figure}
\centering
\input{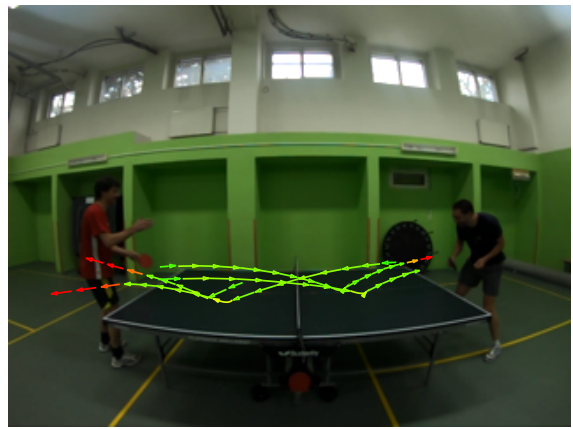}
\caption{Tracking by Deblatting (TbD) successfully recovers trajectory on a pingpong sequence from the proposed TbD dataset. Color encodes Trajectory Intersection over Union (TIoU) with ground truth trajectories from high-speed camera. Arrows indicate the direction of the motion.}
\label{fig:teaser}
\end{figure}


Standard trackers, both long and short term, provide information about the object location in a frame in the from of a single rectangle. The true, continuous trajectory of the object center is thus sampled with the frequency equal to the video frame rate. 
For slow moving objects, such sampling is adequate.
For fast moving objects, especially if their trajectory is not linear (due to bounces, gravity, friction), a single location estimate per frame cannot represent the true trajectory well, even if the fast moving object is inside the reported bounding box. Moreover, standard trackers typically fail even in achieving that \cite{fmo}.

We propose a novel methodology for tracking fast-moving,
blurred objects.
The approach untangles the image formation by solving two inverse problems: {\it motion deblurring} and {\it image matting}.
We therefore call the method   \emph{Tracking by Deblatting}, TbD in short. 

The deblatting procedure simultaneously recovers the trajectory of the object, its shape and appearance. 
We introduce a strong prior on the blur kernel and force it to lie on a 1D manifold.   
The corresponding curve models the object trajectory within a frame.
Unlike a standard general tracker, TbD does not need a template of the object, since the representation of the shape and appearance of the object is recovered on the fly.
Experiments show that the estimated trajectory is often highly accurate; see Fig.~\ref{fig:teaser}.

\section{Related work}
\label{sec:Rel}
 
Object tracking methods are based on diverse principles, such as correlation~\cite{Biresaw:2015:CST:2729305.2741247,dsst,srdcf,csrdcf,Tang2018}, feature point tracking~\cite{tomasi1991detection}, mean-shift~\cite{Comaniciu:2003:KOT:776753.776799,asms}, and tracking-by-detection~\cite{meem,struck}. In addition, several surveys of object tracking have been compiled \cite{Avidan:2007:ET:1191552.1191804,5674053,Godec:2013:HTN:2527401.2527616}. Excellent performance in visual object tracking has been shown by discriminative correlation filters~\cite{Biresaw:2015:CST:2729305.2741247,dsst,srdcf,csrdcf}, yet all the methods fail when the tracked object is blurred as demonstrated in \cite{fmo}. 


Methods proposed for object motion deblurring try to estimate sharp images from photos or videos without considering the tracking goal. Early methods worked with a transparency map (the alpha matte) caused by the blur, and assumed linear motion \cite{Jia2007,daiwu:08} or rotation \cite{Shan2007}. Blind deconvolution of the transparency map is better posed, since the latent sharp map is a binary image.
Accurate estimation of the transparency map by alpha matting algorithms, such as \cite{Levin2008}, is necessary and this is not tractable for large blurs.
Other methods are based on the observation that autocorrelation increases in the direction of blur \cite{Kim2014,Sun2015}. Autocorrelation techniques require a relatively large neighborhood to estimate blur parameters and such methods are not suitable for small moving objects.
More recently, deep learning has been applied to motion deblurring of videos \cite{Wieschollek2017,Su2017} and to the generation of intermediate short-exposure frames \cite{Jin2018}. The proposed convolutional neural networks are trained only on small blurs; blur parameters are not available as they are not directly estimated. 

Tracking methods that consider motion blur has been proposed in \cite{Wu2011,Seibold2017,Ma2016}, yet there is an important distinction between models therein and the FMO problem considered here. The blur is assumed to be caused by camera motion and not by the object motion, which results in blur affecting the whole image and in the absence of alpha blending of the tracked object with the background.

To our knowledge, the only method that tackles the similar problem of tracking motion-blurred objects remains the work in \cite{fmo}. The authors assume linear motion and the trajectories are calculated by fitting a line segment to a morphologically thinned difference image between the given frame and the estimated background.

\section{Tracking By Deblatting}
\label{sec:TbD_overview}

The proposed method formulates tracking as an inverse problem to the video formation model. Suppose that within a single video frame $I$ an object $F$ moves along the trajectory $\C$ in front of background $B$. Frame $I$ is then formed as
\begin{equation}
	\label{eq:acquisition_model}
	I = H*F + (1-H*M)B,
\end{equation}
where $*$ denotes convolution, $H$ is the Point Spread Function (PSF) of the object motion blur corresponding to trajectory $\C$, and $M$ is the binary mask of the object shape (\ie the indicator function of $F$). We refer to the pair $(F, M)$  as the object model. 
The first term is the tracked object blurred by its own motion, the second term is the background partially occluded by the object, and the blending coefficients are determined by $H*M$.
Inference under the assumption of this formation model consists of solving simultaneously two inverse problems:
blind deblurring and image matting. The solution is the estimated PSF $H$ and the object model $F$ and $M$.

Motion blur in \eqref{eq:acquisition_model} is modeled by convolution, which implies the following assumption about the object motion: The object shape and appearance remain constant during the frame exposure time. Scenarios that satisfy the assumption precisely are, \eg, an object of arbitrary shape undergoing only translational motion or a spherical object of uniform color undergoing arbitrary motion under spatially-uniform illumination. In addition, the motion must be in a plane parallel to the camera image plane to guarantee constant size of the object. For the purpose of tracking and trajectory estimation we claim that the formation model \eqref{eq:acquisition_model} with convolution is sufficient as long as the assumption holds at least approximately, which is experimentally validated on the presented dataset. 



\begin{figure}[!t]
\centering
\includegraphics[width=0.99\linewidth]{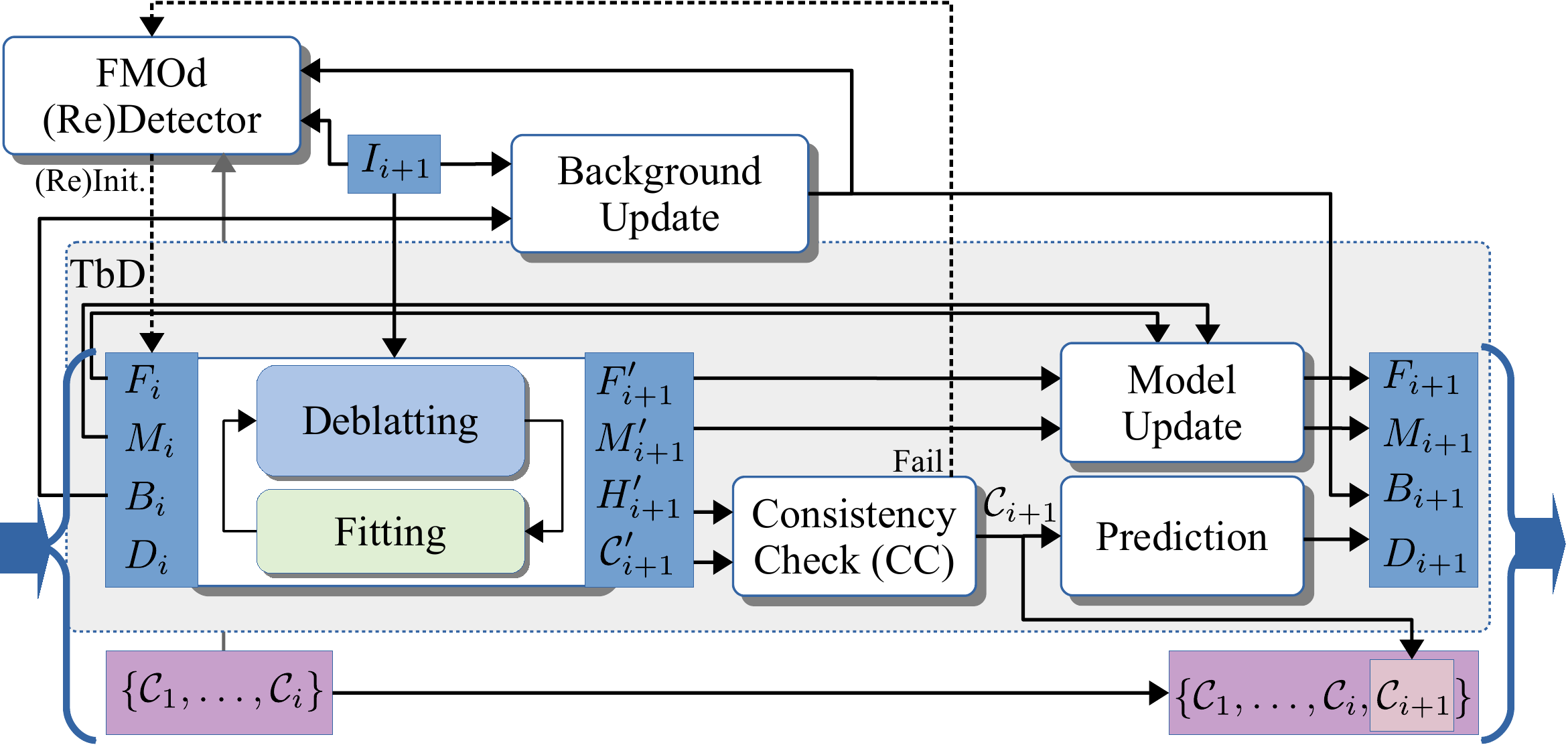}
\caption{Long-term Tracking by Deblatting (Sec.~\ref{sec:TbD_overview}). The FMO detector is activated during initialization or if the consistency check fails.
\label{fig:pipeline_whole}}
\end{figure}

\begin{figure}[!t]
\centering
\includegraphics[width=0.99\linewidth]{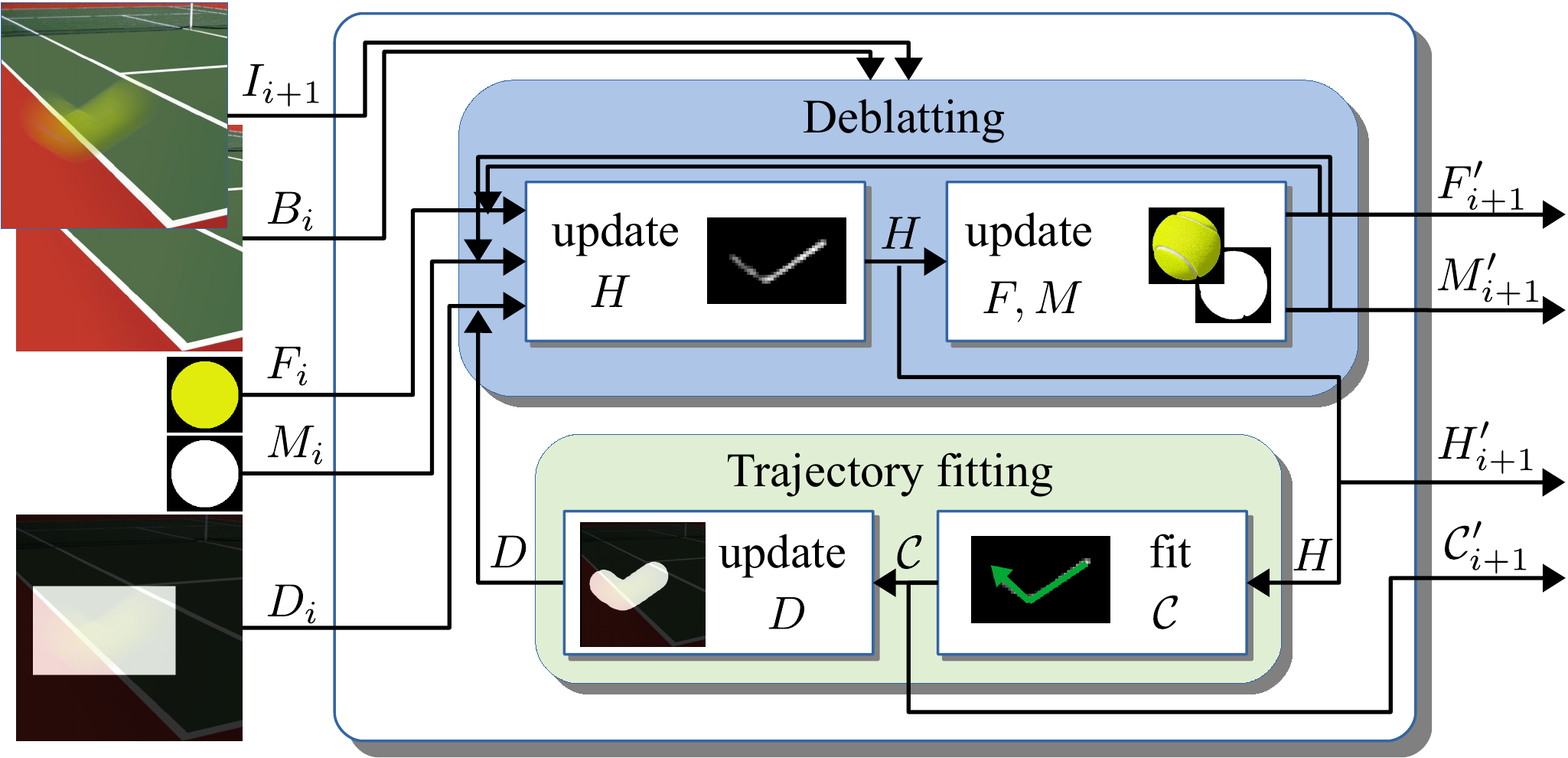}
\caption{{\it Deblatting}, \ie~{\it debl}urring and m{\it atting} -- Sec.~\ref{sec:deblurring}, with trajectory fitting -- Sec.~\ref{sec:fitting}. 
\label{fig:pipeline_deblatting}}
\end{figure}

The proposed method is iterative and causal processing of a new frame $I_{i+1}$ using only knowledge acquired from earlier frames ${I_1,\ldots,I_i}$; Fig. \ref{fig:pipeline_whole} (shaded area) provides the overview. Inputs are the current estimates of the object model $F_i$ and $M_i$, background $B_i$, and a region of interest $D_i$ in $I_{i+1}$, which is the neighborhood of the predicted object location. Three main steps are performed in TbD:
\vspace*{-0.5em}
\begin{enumerate}[leftmargin=*]
\setlength\itemsep{-0.25em}
	\item \emph{Deblatting}: Iteratively solve blind deblurring and matting in the image region $D_i$ with the model \eqref{eq:acquisition_model} and estimate $F'_{i+1}$, $M'_{i+1}$, and $H_{i+1}$; see Sec.~\ref{sec:deblurring}
	\item \emph{Trajectory fitting}: Estimate physically plausible motion trajectory (parametric curve) $\C_{i+1}$ corresponding to $H_{i+1}$ and optionally adjust $D_i$ according to $\C_{i+1}$; see Sec.~\ref{sec:fitting}.
	\item \emph{Consistency check \& model update}: Verify that the error of the mapping $H\to \C$ is below threshold $\tau$, predict the new region of interest $D_{i+1}$ for the next frame, and update the object model to $F_{i+1}$ and $M_{i+1}$.
\end{enumerate}
\vspace*{-0.5em}
A more detailed illustration of Steps 1 and 2 is in Fig.~\ref{fig:pipeline_deblatting}. Step 1 stops after reaching either a given relative tolerance or a maximum number of iterations. Steps 1 and 2 are repeated only if the newly fitted $\C$ touches the boundary of $D$~--~in this case the new $D$ is the $d-$neighborhood of $\C$ where $d$ is the object diameter. Adjusting $D$ this way helps to eliminate the detrimental influence of other moving objects to correct estimation of $H$.

If the consistency check (CC) passes, we extrapolate the estimated trajectory to the next frame and $D_{i+1}$ is again $d$-neighborhood of this extrapolation. To update the appearance model we use exponential forgetting
\begin{equation}\label{eq:update_model}
F_{i+1} = \gamma F_{i} + (1-\gamma)F'_{i+1};
\end{equation}
$M$ is updated analogically. 

To enable long-term tracking, the FMO detector (FMOd) from \cite{fmo} determines the new input if CC fails.
First, FMOd tries detecting the object in an gradually enlarged $D$. If it succeeds, the main TbD pipeline is reinitialized with $D$ set as a neighborhood of the FMOd detection. If FMOd fails, TbD returns the extrapolation of trajectory $\C_i$ as the best guess of $\C_{i+1}$ and tracking is restarted anew on the next frame. The background $B_i$ is estimated as a temporal median of frames $B_{i-1},\,B_{i-2},\ldots$, optionally including video stabilization if necessary. The first detection is also performed automatically by FMOd.  The object appearance model is either learned ``on the fly'' starting trivially with $F_0\equiv1,\,M_0\equiv1$, or the user provides a template of the tracked object, \eg~a rectangular region from one of the frames where the object is still.


\subsection{Deblatting}
\label{sec:deblurring}
The core step of TbD is the extraction of motion information $H$ from the input frame, which we formulate as a blind deblurring and matting problem. Inputs are the frame $I$, domain $D$, background $B$, and the object appearance model $\hat F$. The inverse problem corresponding to \eqref{eq:acquisition_model} is formulated as
\begin{equation}
	\label{eq:deconv_loss}
	\begin{gathered}
	\min_{F,M,H} \frac{1}{2}\left\|H*F+(1-H*M)B-I\right\|_2^2 \\
	+\frac{\lambda}{2}\|F-M\hat F\|_2^2+\alpha_F\|\nabla F\|_1+\alpha_H\|H\|_1\\
	\end{gathered}
\end{equation}
s.t. $0\leq F\leq M\leq 1$ and $H\geq 0$ in $D$, $H\equiv 0$ elsewhere.
The primary unknown is $H$, but $F$ and $M$ are estimated as by-products. The first term in \eqref{eq:deconv_loss} is the fidelity to the model \eqref{eq:acquisition_model}. The second $\lambda$-weighted term is a form of ``template-matching'', an agreement with a prescribed appearance. The template $\hat{F}$ is multiplied by $M$ because if $\hat F$ is initially supplied by user as a rectangular region from a video frame, it contains the object and partially also the surrounding background. 
When processing the $i$-th frame,
we set $\hat F=F_{i-1}$ as the updated appearance estimate \eqref{eq:update_model} from the previous frame. The first $L^1$ term is the total variation that promotes smoothness of the recovered object appearance. The second $L^1$ regularization enforces sparsity of the blur and reduces small nonzero values.

If $M$ is a binary mask then the condition $F\leq M$ states that $F$ cannot be nonzero where $M$ is zero -- pixels outside the object must be zero. For computational reasons, we relax the binary restriction and allow $M$ to attain values in the range $[0,1]$. The correct constraint corresponding to this relaxation is then exactly $F\leq M$, assuming $F$ alone is bounded in $[0,1]$. The inequality constraint $H\geq 0$ prohibits negative values in $H$, which are physically implausible for motion blur, and $H$ is estimated only within the domain $D$.

We solve \eqref{eq:deconv_loss} in an alternating manner, fix $(F,M)$ and solve for $H$ and vice versa,
until convergence.

Minimizing \eqref{eq:deconv_loss} w.r.t. $H$ with $(F,M)$ fixed becomes
\begin{equation}
	\label{eq:deconv_H_loss}
	\min_H\frac{1}{2}\left\|H*F+(1-H*M)B-I\right\|_2^2+\alpha_H\|H\|_1
\end{equation}
s.t. $H \geq 0$. We use ADMM to solve \eqref{eq:deconv_H_loss}, which leads to the linear system
\begin{multline}
	\label{eq:deconv_H_eq}
	\left((\mathbf{F}-\mathbf{B}\mathbf{M})^T(\mathbf{F}-\mathbf{B}\mathbf{M})+\rho\right)H \\
	=(\mathbf{F}-\mathbf{B}\mathbf{M})^T(I-B)+\rho(z-u),
\end{multline}
where $\mathbf{F}$ and $\mathbf{M}$ are the convolution operator given by $F$ (\ie~convolution with $F$) and $M$, respectively. $\mathbf{B}$ is the pixelwise multiplication by background $B$ and $z,u,\rho$ are related to ADMM variable splitting for the non-smooth $L^1$ term and the inequality constraint.

Minimizing \eqref{eq:deconv_loss} w.r.t. the joint unknown $(F,M)$ with $H$ fixed is
\begin{multline}
	\label{eq:deconv_F_loss}
	\min_{F,M}\frac{1}{2}\left\|H*F+(1-H*M)B-I\right\|_2^2\\
	+\frac{\lambda}{2}\|F-M\hat F\|_2^2	+\alpha_F\|\nabla F\|_1
\end{multline}
s.t. $0\leq F\leq M\leq 1$.
We again solve this problem using ADMM, which leads to the linear system
\begin{equation}
	\label{eq:deconv_F_eq}
	\begin{gathered}
	\begin{bmatrix} \mathbf{H}^T\mathbf{H}+\rho_1\nabla^T\nabla+\lambda+\rho_2 & \hspace{-1em}-\mathbf{H}^T\mathbf{B}-\lambda\hat F \\ -\mathbf{H}^T\mathbf{B}-\lambda\hat F & \hspace{-1em}-\mathbf{H}^T\mathbf{B}^2\mathbf{H}+\lambda\hat F^2 + \rho_2 \end{bmatrix} \begin{bmatrix} F \\ M \end{bmatrix} \\
	= [\mathbf{H},\, -\mathbf{BH} ]^T (I-B)	+ \rho_1 \nabla^T (z_1-u_1) + \rho_2(z_2-u_2),%
	\end{gathered}
\end{equation}
where $\mathbf{H}$ is the convolution operator given by $H$, and $z_1$, $u_1$, $\rho_1$ are related to ADMM variable splitting due to the nonsmooth regularization. To enforce the constraint $(F,M)\in C$ where $C$ is a convex set defined by $0\leq F\leq M\leq 1$, we use the ADMM splitting $z_2:=(F,M)$ and then each ADMM iteration requires projecting $z_2$ onto $C$. Note that $C\subset\mathbb{R}^4$ and correspondingly $z_2\in\mathbb{R}^4$ since each pixel in $F$ has three (RGB) channels and $M$ is a single-channel mask. Since $C$ is an intersection of half-spaces, we can use iterative Dykstra's projection algorithm \cite{dykstra}. The rest of the minimization is standard.

To summarize, the alternating $H$--$(F,M)$ estimation loop for the $i$-th frame proceeds as follows:
\vspace*{-0.5em}
\begin{enumerate}[leftmargin=*]
	\setlength\itemsep{-0.25em}
	\item Initialize $M:=M^{i-1}$ (if available from previous detection) or $M\equiv 1$; initialize $\hat F:=F^{i-1}$, $F:=M\hat F$.
	\item Calculate $H$ by solving \eqref{eq:deconv_H_loss}.
	\item Check convergence, exit if satisfied.
	\item Calculate $(F,M)$ by solving \eqref{eq:deconv_F_loss}, go to 2.
\end{enumerate}

\subsection{Trajectory fitting}
\label{sec:fitting}

Fitting the PSF $H$, which is a gray-scale image, with a trajectory $\C(t): [0,1]\to\mathbb{R}^2$ serves three purposes. First, we use the error of the fit in the Consistency Check to determine if $H$
is the motion blur induced by the tracked object and thus whether to proceed with tracking,
or to declare the deblatting step a failure and 
to reinitialize it with different parameters. Second, the trajectory as an analytic curve can be used for motion prediction whereas $H$ cannot. Third, $\C$ defines the intra-frame motion, which is the desired output of the proposed method.


The fitting is analogous to vectorization of raster images. It is formulated as the maximum a posteriori estimation of $\C$, given $H$, with the physical plausibility of the trajectory used as a prior. Let $\mathcal{C}$ be a curve defined by a set of parameters $\theta$ (\eg~polynomial coefficients) and $H_{\C}$ be a raster image of the corresponding $\C$ (\ie~blur PSF). We say that the curve $\C$ is the \emph{trajectory fit} of $H$ if $\theta$ minimizes
\begin{equation}
	\label{eq:fitting_loss_H}
	\min_{\theta} \|H_\mathcal{C}-H\| \quad\text{s.t. }\mathcal{C}\in\Psi,
\end{equation}
where $\Psi$ is the set of admissible curves. 

\begin{figure}
	\noindent\begin{minipage}[t]{.33\linewidth}
		\centering
		\includegraphics[width=\textwidth]{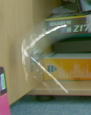}
	\end{minipage}\hfill%
	\begin{minipage}[t]{.33\linewidth}
		\centering
		\includegraphics[width=\textwidth]{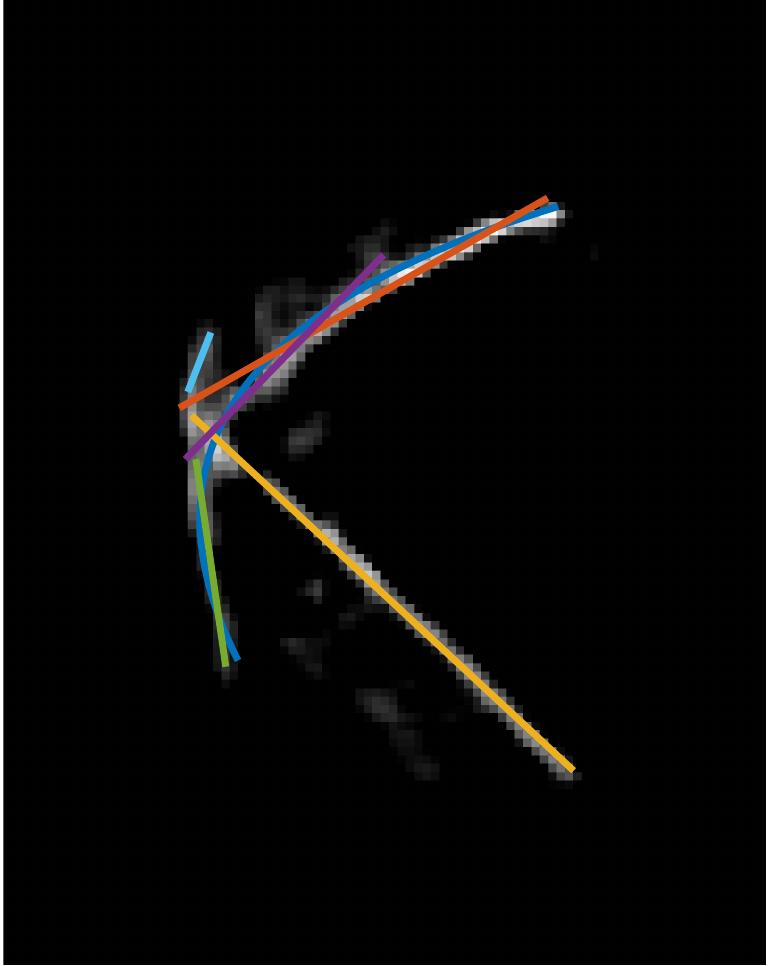}
	\end{minipage}\hfill%
	\begin{minipage}[t]{.33\linewidth}
		\centering
		\includegraphics[width=\textwidth]{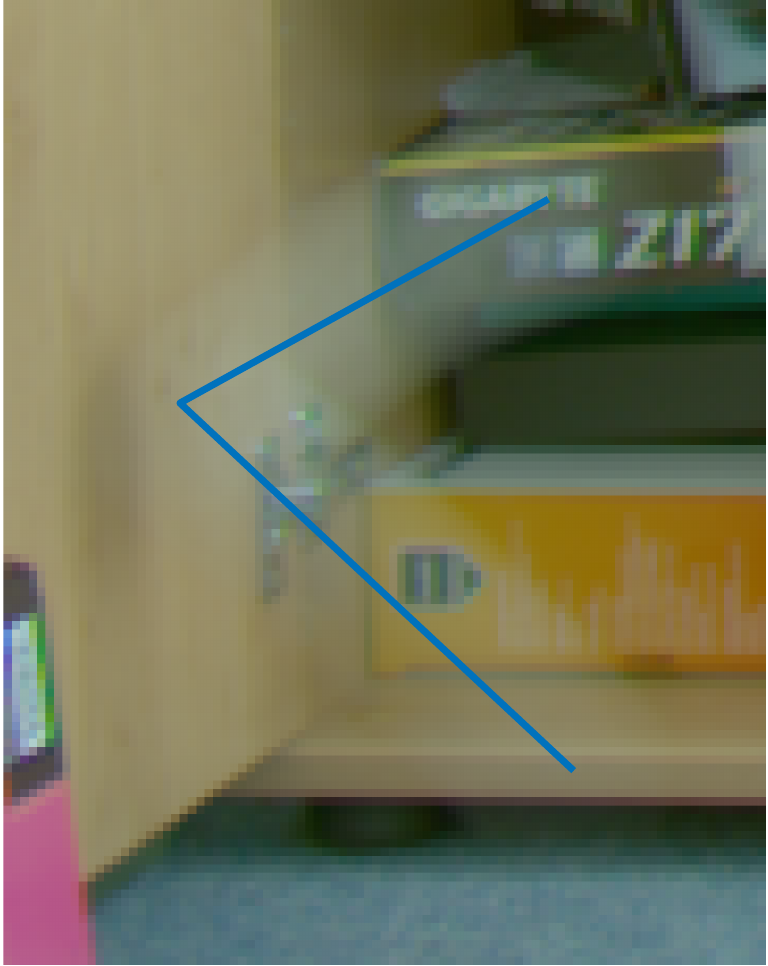}
	\end{minipage}
	\begin{minipage}[t]{.33\linewidth}
		\centering
		\includegraphics[width=\textwidth]{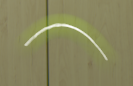}
	\end{minipage}\hfill%
	\begin{minipage}[t]{.33\linewidth}
		\centering
		\includegraphics[width=\textwidth]{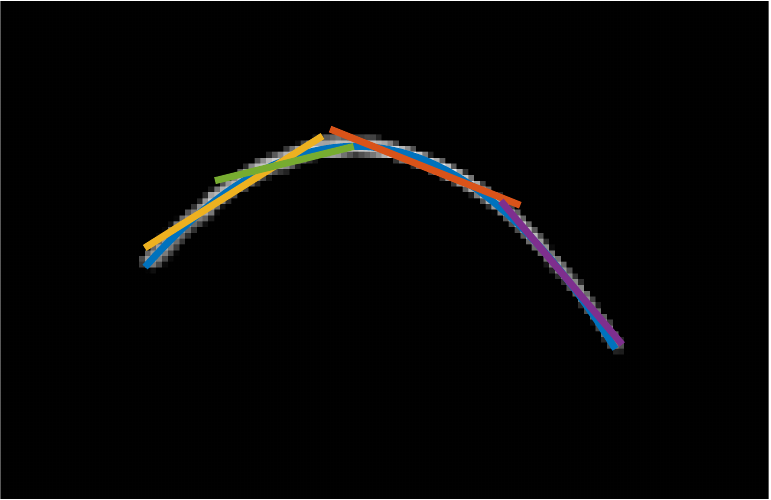}
	\end{minipage}\hfill%
	\begin{minipage}[t]{.33\linewidth}
		\centering
		\includegraphics[width=\textwidth]{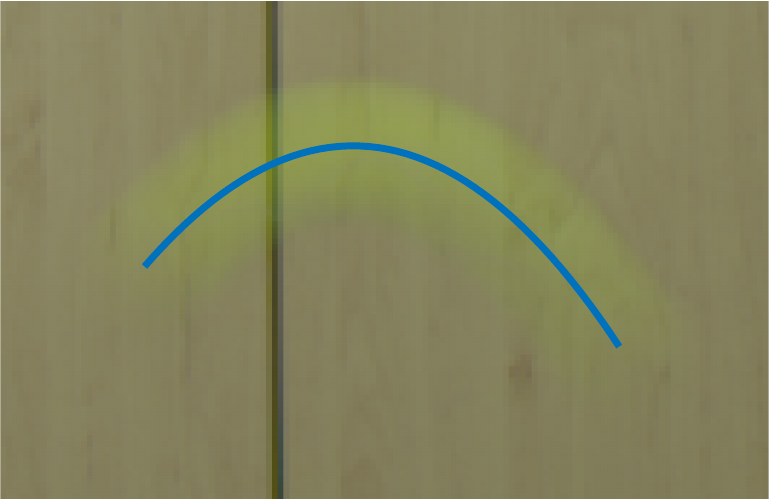}
	\end{minipage}
	\begin{minipage}[t]{.33\linewidth}
		\centering
		$I$ and $H$
	\end{minipage}\hfill%
	\begin{minipage}[t]{.33\linewidth}
		\centering
		RANSAC
	\end{minipage}\hfill%
	\begin{minipage}[t]{.33\linewidth}
		\centering
		$I$ and $\mathcal{C}$
	\end{minipage}
	\caption{Trajectory fitting. \emph{Left} input image with~estima\-ted blur superimposed in white, \emph{middle} linear and parabolic segments found by RANSAC, \emph{right} final fitted \mbox{trajectory.}}
	\label{fig:fitting}
\end{figure}

Our main tracking targets are balls and similar free-falling objects, therefore our assumption is that between impulses from other moving objects (\eg~players), tracked objects remain in free flight or bounce off static rigid bodies. We then define $\Psi$ as a set of piecewise quadratic continuous curves~--~quadratic to account for deacceleration due to gravity and piecewise to account for abrupt change of motion during bounces. $\mathcal{C}\in\Psi$ is defined as
\begin{equation}
	\label{eq:fitting_curve_def}
	\mathcal{C}(t) = \begin{cases} \sum_k^2 c_{k,1}t^k & 0\leq t \leq \tilde{t}, \\
		\sum_k^2 c_{k,2}t^k & \tilde{t} \leq t \leq 1,
	\end{cases}
\end{equation}
s.t. $\sum_k^2 c_{k,1}{\tilde{t}}^k = \sum_k^2 c_{k,2}{\tilde{t}}^k$. Single linear or quadratic curves are included as special cases when $\tilde{t}=1$. 
The problem \eqref{eq:fitting_loss_H} is non-convex and thus a good initial guess is necessary for gradient-descent optimization to perform well. To this end, we employed a four-step procedure:
\vspace*{-0.5em}
\begin{enumerate}[leftmargin=*]
	\setlength\itemsep{-0.25em}
	\item Identify the most salient linear and quadratic segments in $H$ by RANSAC.
	\item Connect segments to form a curve $\mathcal{C}$ of the kind \eqref{eq:fitting_curve_def}.
	\item Refine $\mathcal{C}$ to be a locally optimal fit of $H$ in terms of pointwise distance.
	\item Calculate the loss \eqref{eq:fitting_loss_H} and choose the best candidate.
\end{enumerate}
\vspace*{-0.5em}
See Fig.~\ref{fig:fitting} for illustrations of the above steps. 

Let us view the blur $H$ as a set of pixels with coordinates $x_i$ and intensities $w_i>0$. Sequential RANSAC finds line segments as follows: sample two points, find inliers of the corresponding line, find the most salient consecutive run of points on this line and in each round remove the winner from the sampling pool.
The saliency is defined as $\sum w_i$ for $x_i$ in the inlier set and ``consecutive'' means that the distance between neighboring points is bounded by a threshold.
The search stops when the saliency drops bellow a specified threshold or there are no more points.
We denote the set of collected linear segments as $\mathcal{M}_1$. Parabolic arcs are found similarly. We sample four points, find two corresponding parabolas, project the remaining points on the parabolas to determine the distance and inlier set as well as the arc-length parametrization of inliers (required for correct ordering and mutual distance calculation of inliers) and again find the most salient consecutive run. 
We denote the set of collected parabolic segments as $\mathcal{M}_2$.

The solution will be close to a curve formed from one or two segments (linear or parabolic) found so far. Let $\C_1, \C_2 \in \mathcal{M}_1$ be two linear segments. If the intersection $P$ of the corresponding lines is close to the segments (w.r.t. some threshold), the curve connecting $\mathcal{C}_1\to P\to\mathcal{C}_2$ is a candidate for the piecewise linear trajectory fit. This way we construct a set $\mathcal{M}_3$ of all candidate and similarly $\mathcal{M}_4$ with candidates of parabolic pairs.

Curves in $\mathcal{M}_0=\bigcup\mathcal{M}_i$ are approximate candidates for the final trajectory, yet we first refine them to be locally optimal robust fits to $H$. We say that a curve $\mathcal{C}$ defined by a set of parameters $\theta$ is locally optimal fit to $\{x_i\}$ if $\theta$ is the minimizer of the problem
\begin{equation}
	\label{eq:fitting_loss_dist}
	\min_\theta \sum_{x_i\in K} w_i\operatorname{dist}(x_i,\,\mathcal{C}) + \lambda\int_0^1\operatorname{dist}(\mathcal{C}(t), \{x_i\})\mathrm{d}t\,
\end{equation}
where $K=\{x_i|\,\operatorname{dist}(x_i,\,\mathcal{C})<\rho\}$, $\operatorname{dist}(x, \C)$ is the distance of the point $x$ to the curve $\C$ and $\operatorname{dist}(\C(t), \{x_i\})$ is the distance of the curve point $C(t)$ to the set $\{x_i\}$. In the first term, $K$ is a set of inliers defined by the distance threshold $\rho$ and then $\C$ is the distance-optimized fit to inliers. The second term restricts curve length.

The gradient of \eqref{eq:fitting_loss_dist} is intractable since the distance of a point $x$ to a non-convex set (in our case the curve $\C$) is intractable. We therefore resort to a procedure similar to the Iterative Closest Point (ICP) algorithm. In each iteration, we fix the currently closest curve counterpart $y_i=\C(t_i)$ for each point $x_i$ by solving $t_i = \operatorname{argmin}_t\operatorname{dist}(x_i,\mathcal{C}(t))$, and in \eqref{eq:fitting_loss_dist} we approximate $\operatorname{dist}(x_i,\mathcal{C})\approx\|x_i-y_i\|$ and analogically for $\operatorname{dist}(\mathcal{C}(t), \{x\})$. 
Then \eqref{eq:fitting_loss_dist} becomes a tractable function of $\theta$. We find the solution using the Iteratively Reweighted Least Squares algorithm and proceed with the next iteration of ICP.
The algorithms converges in a few iterations and the optimization is fast.

We then refine every curve $\mathcal{C}_0\in\mathcal{M}_0$ by solving \eqref{eq:fitting_loss_dist} with the ICP-like algorithm and denote the set of solutions as $\mathcal{M}$. 
Finally, for each curve $\mathcal{C}\in\mathcal{M}$ we construct $H_\mathcal{C}$, measure the error $\|H_\mathcal{C}-H\|$ and choose the best candidate as the trajectory fit. 
In TbD, the Consistency Check of the trajectory fit $\C$ is performed by evaluating the criterion $\|H_\mathcal{C}-H\|/\|H\|<\tau$.

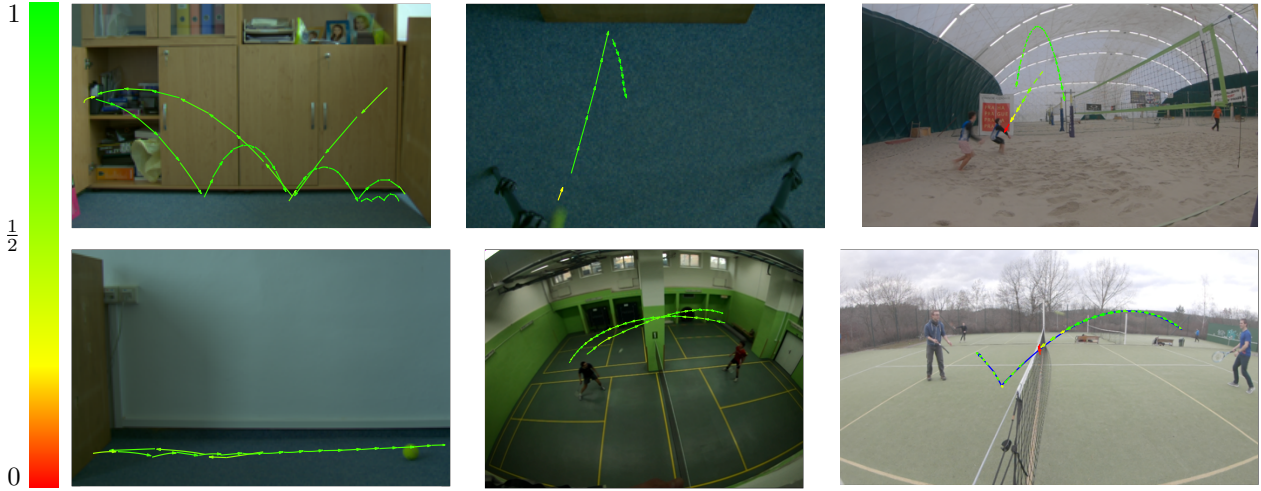
\begin{figure*}
\centering
\begin{tabular}{@{}c@{}c@{}}
\raisebox{-.51\height}[0pt][0pt]{ 
\begin{tabular}{@{}r@{}}
\raisebox{25.5\height}[0pt][0pt]{1} \\
\raisebox{11\height}[0pt][0pt]{$\frac{1}{2}$}  \\
\raisebox{2\height}[0pt][0pt]{0} \\ 
\end{tabular}
\includegraphics[height=0.37\textwidth]{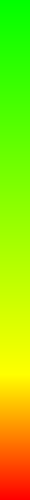}}
 & \resizebox {!}{0.18\textwidth} {\begin{tikzpicture} 
\begin{axis}[y dir=reverse, 
 xmin=1,xmax=960, 
 ymin=1,ymax=600, 
 xticklabels = \empty, yticklabels = \empty, 
 grid=none, axis equal image] 
\addplot graphics[xmin=1,xmax=960,ymin=1,ymax=600] {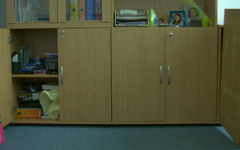}; 
\addplot [<-,>={Latex[length=1.5mm,width=0.5mm,angle'=25,open,round]},,domain=0:1,samples=2,style=semithick,color={rgb,255:red,102; green,255; blue,0}]({776.0582 + 44.0527*x},{294.056 + -45.4301*x});  
\addplot [>={Latex[length=1.5mm,width=0.5mm,angle'=25,open,round]},,domain=0:1,samples=2,style=semithick,color={rgb,255:red,102; green,255; blue,0}]({820.1109 + 25.5333*x},{248.6258 + -24.2483*x});  
\addplot [<-,>={Latex[length=1.5mm,width=0.5mm,angle'=25,open,round]},,domain=0:1,samples=2,style=semithick,color={rgb,255:red,56; green,255; blue,0}]({679.7763 + 84.6664*x},{400.9473 + -98.4267*x});  
\addplot [->,>={Latex[length=1.5mm,width=0.5mm,angle'=25,open,round]},,domain=0:1,samples=10,style=semithick,color={rgb,255:red,54; green,255; blue,0}]({675.8328 + -73.8044*x + -6.1623*x^2},{407.4982 + 97.1571*x + 11.1915*x^2});  
\addplot [->,>={Latex[length=1.5mm,width=0.5mm,angle'=25,open,round]},,domain=0:1,samples=2,style=semithick,color={rgb,255:red,84; green,255; blue,0}]({590.2251 + -70.4604*x},{515.1294 + -86.7505*x});  
\addplot [->,>={Latex[length=1.5mm,width=0.5mm,angle'=25,open,round]},,domain=0:1,samples=10,style=semithick,color={rgb,255:red,64; green,255; blue,0}]({514.1435 + -60.6585*x + -12.146*x^2},{423.324 + -68.9468*x + -2.2594*x^2});  
\addplot [<-,>={Latex[length=1.5mm,width=0.5mm,angle'=25,open,round]},,domain=0:1,samples=2,style=semithick,color={rgb,255:red,61; green,255; blue,0}]({367.9831 + 72.6916*x},{294.8566 + 54.5788*x});  
\addplot [<-,>={Latex[length=1.5mm,width=0.5mm,angle'=25,open,round]},,domain=0:1,samples=2,style=semithick,color={rgb,255:red,77; green,255; blue,0}]({291.4055 + 75.571*x},{254.2222 + 39.4049*x});  
\addplot [<-,>={Latex[length=1.5mm,width=0.5mm,angle'=25,open,round]},,domain=0:1,samples=2,style=semithick,color={rgb,255:red,42; green,255; blue,0}]({214.4451 + 74.0091*x},{232.7138 + 19.019*x});  
\addplot [<-,>={Latex[length=1.5mm,width=0.5mm,angle'=25,open,round]},,domain=0:1,samples=2,style=semithick,color={rgb,255:red,80; green,255; blue,0}]({139.6245 + 77.9128*x},{227.3486 + 4.5071*x});  
\addplot [<-,>={Latex[length=1.5mm,width=0.5mm,angle'=25,open,round]},,domain=0:1,samples=10,style=semithick,color={rgb,255:red,66; green,255; blue,0}]({69.1721 + 75.5332*x + -5.1982*x^2},{242.0404 + -23.8772*x + 11.5855*x^2});  
\addplot [->,>={Latex[length=1.5mm,width=0.5mm,angle'=25,open,round]},,domain=0:1,samples=10,style=semithick,color={rgb,255:red,167; green,255; blue,0}]({32.9601 + -5.4732*x + 35.9281*x^2},{265.8294 + -25.0221*x + 7.9521*x^2});  
\addplot [->,>={Latex[length=1.5mm,width=0.5mm,angle'=25,open,round]},,domain=0:1,samples=2,style=semithick,color={rgb,255:red,23; green,255; blue,0}]({64.3134 + 51.6755*x},{255.6108 + 15.7891*x});  
\addplot [->,>={Latex[length=1.5mm,width=0.5mm,angle'=25,open,round]},,domain=0:1,samples=10,style=semithick,color={rgb,255:red,51; green,255; blue,0}]({116.7092 + 62.788*x + -5.6605*x^2},{272.2472 + 24.105*x + 9.8399*x^2});  
\addplot [->,>={Latex[length=1.5mm,width=0.5mm,angle'=25,open,round]},,domain=0:1,samples=2,style=semithick,color={rgb,255:red,50; green,255; blue,0}]({175.0955 + 55.7575*x},{304.9729 + 47.699*x});  
\addplot [<-,>={Latex[length=1.5mm,width=0.5mm,angle'=25,open,round]},,domain=0:1,samples=2,style=semithick,color={rgb,255:red,47; green,255; blue,0}]({288.425 + -55.9342*x},{421.1377 + -65.6299*x});  
\addplot [->,>={Latex[length=1.5mm,width=0.5mm,angle'=25,open,round]},,domain=0:1,samples=10,style=semithick,color={rgb,255:red,66; green,255; blue,0}]({289.9032 + 63.6241*x + -5.8659*x^2},{425.695 + 82.8388*x + 1.6747*x^2});  
\addplot [->,>={Latex[length=1.5mm,width=0.5mm,angle'=25,open,round]},,domain=0:1,samples=2,style=semithick,color={rgb,255:red,56; green,255; blue,0}]({354.7441 + 23.3464*x},{518.5538 + -49.7157*x});  
\addplot [->,>={Latex[length=1.5mm,width=0.5mm,angle'=25,open,round]},,domain=0:1,samples=2,style=semithick,color={rgb,255:red,32; green,255; blue,0}]({377.3389 + 27.0497*x},{465.5113 + -44.3839*x});  
\addplot [->,>={Latex[length=1.2736mm,width=0.5mm,angle'=25,open,round]},,domain=0:1,samples=2,style=semithick,color={rgb,255:red,24; green,255; blue,0}]({404.4613 + 27.3838*x},{417.9464 + -26.647*x});  
\addplot [->,>={Latex[length=1.0497mm,width=0.5mm,angle'=25,open,round]},,domain=0:1,samples=2,style=semithick,color={rgb,255:red,32; green,255; blue,0}]({431.4289 + 29.1388*x},{390.1063 + -11.9413*x});  
\addplot [->,>={Latex[length=0.99547mm,width=0.5mm,angle'=25,open,round]},,domain=0:1,samples=10,style=semithick,color={rgb,255:red,25; green,255; blue,0}]({461.388 + 29.0111*x + -1.4201*x^2},{380.6301 + -11.0551*x + 17.083*x^2});  
\addplot [->,>={Latex[length=1.1711mm,width=0.5mm,angle'=25,open,round]},,domain=0:1,samples=2,style=semithick,color={rgb,255:red,28; green,255; blue,0}]({491.0177 + 27.5984*x},{383.3429 + 21.7403*x});  
\addplot [->,>={Latex[length=1.5mm,width=0.5mm,angle'=25,open,round]},,domain=0:1,samples=10,style=semithick,color={rgb,255:red,36; green,255; blue,0}]({517.0831 + 31.8231*x + -3.4954*x^2},{406.9382 + 31.9741*x + 6.9111*x^2});  
\addplot [<-,>={Latex[length=1.5mm,width=0.5mm,angle'=25,open,round]},,domain=0:1,samples=2,style=semithick,color={rgb,255:red,43; green,255; blue,0}]({574.4907 + -27.7103*x},{504.7709 + -56.6525*x});  
\addplot [->,>={Latex[length=1.2064mm,width=0.5mm,angle'=25,open,round]},,domain=0:1,samples=2,style=semithick,color={rgb,255:red,118; green,255; blue,0}]({581.7963 + 20.0689*x},{528.8638 + -30.1198*x});  
\addplot [->,>={Latex[length=1.4603mm,width=0.5mm,angle'=25,open,round]},,domain=0:1,samples=2,style=semithick,color={rgb,255:red,40; green,255; blue,0}]({602.2376 + 26.7274*x},{495.9129 + -34.7098*x});  
\addplot [->,>={Latex[length=1.094mm,width=0.5mm,angle'=25,open,round]},,domain=0:1,samples=10,style=semithick,color={rgb,255:red,23; green,255; blue,0}]({631.5186 + 22.1237*x + 4.7468*x^2},{458.9947 + -24.8987*x + 6.6514*x^2});  
\addplot [<-,>={Latex[length=0.90661mm,width=0.5mm,angle'=25,open,round]},,domain=0:1,samples=10,style=semithick,color={rgb,255:red,38; green,255; blue,0}]({687.3443 + -27.5663*x + 1.6298*x^2},{439.8967 + -11.24*x + 13.6124*x^2});  
\addplot [->,>={Latex[length=1.0405mm,width=0.5mm,angle'=25,open,round]},,domain=0:1,samples=10,style=semithick,color={rgb,255:red,26; green,255; blue,0}]({688.4273 + 29.153*x + -2.057*x^2},{440.1391 + 3.6215*x + 10.6735*x^2});  
\addplot [<-,>={Latex[length=1.4448mm,width=0.5mm,angle'=25,open,round]},,domain=0:1,samples=2,style=semithick,color={rgb,255:red,34; green,255; blue,0}]({743.1773 + -27.1328*x},{488.1604 + -33.8015*x});  
\addplot [->,>={Latex[length=1.5mm,width=0.5mm,angle'=25,open,round]},,domain=0:1,samples=10,style=semithick,color={rgb,255:red,38; green,255; blue,0}]({742.1443 + 28.2507*x + -3.5443*x^2},{488.2116 + 40.7333*x + 1.1249*x^2});  
\addplot [->,>={Latex[length=1.4701mm,width=0.5mm,angle'=25,open,round]},,domain=0:1,samples=10,style=semithick,color={rgb,255:red,38; green,255; blue,0}]({768.8747 + 27.6002*x + 1.2421*x^2},{523.0746 + -41.1803*x + 8.0097*x^2});  
\addplot [->,>={Latex[length=1.0547mm,width=0.5mm,angle'=25,open,round]},,domain=0:1,samples=2,style=semithick,color={rgb,255:red,16; green,255; blue,0}]({797.7451 + 26.9564*x},{488.5852 + -16.5708*x});  
\addplot [->,>={Latex[length=0.90009mm,width=0.5mm,angle'=25,open,round]},,domain=0:1,samples=2,style=semithick,color={rgb,255:red,38; green,255; blue,0}]({826.5003 + 27.0028*x},{472.3493 + 0.052249*x});  
\addplot [->,>={Latex[length=1.0404mm,width=0.5mm,angle'=25,open,round]},,domain=0:1,samples=2,style=semithick,color={rgb,255:red,42; green,255; blue,0}]({853.9424 + 25.5757*x},{474.8676 + 17.8917*x});  
\addplot [<-,>={Latex[length=0.77978mm,width=0.5mm,angle'=25,open,round]},,domain=0:1,samples=2,style=semithick,color={rgb,255:red,106; green,255; blue,0}]({893.4655 + -13.9378*x},{510.7836 + -18.7879*x});  
\addplot [<-,>={Latex[length=0.64405mm,width=0.5mm,angle'=25,open,round]},,domain=0:1,samples=10,style=semithick,color={rgb,255:red,53; green,255; blue,0}]({866.1004 + 14.537*x + -3.9133*x^2},{513.8838 + 12.5971*x + 3.2755*x^2});  
\addplot [<-,>={Latex[length=0.5mm,width=0.5mm,angle'=25,open,round]},,domain=0:1,samples=2,style=semithick,color={rgb,255:red,25; green,255; blue,0}]({852.0439 + 13.8413*x},{509.7853 + 2.8312*x});  
\addplot [<-,>={Latex[length=0.72703mm,width=0.5mm,angle'=25,open,round]},,domain=0:1,samples=10,style=semithick,color={rgb,255:red,32; green,255; blue,0}]({836.5888 + 10.312*x + 3.9792*x^2},{527.6812 + -19.7497*x + 3.5624*x^2});  
\addplot [<-,>={Latex[length=0.65036mm,width=0.5mm,angle'=25,open,round]},,domain=0:1,samples=2,style=semithick,color={rgb,255:red,33; green,255; blue,0}]({822.7899 + 15.6978*x},{516.43 + 11.5867*x});  
\addplot [<-,>={Latex[length=0.63369mm,width=0.5mm,angle'=25,open,round]},,domain=0:1,samples=10,style=semithick,color={rgb,255:red,30; green,255; blue,0}]({805.1248 + 14.9144*x + 1.4033*x^2},{525.92 + -11.8913*x + 2.2586*x^2});  
\addplot [<-,>={Latex[length=0.55863mm,width=0.5mm,angle'=25,open,round]},,domain=0:1,samples=2,style=semithick,color={rgb,255:red,35; green,255; blue,0}]({790.552 + 14.702*x},{521.9049 + 8.0447*x});  
\addplot [<-,>={Latex[length=0.5799mm,width=0.5mm,angle'=25,open,round]},,domain=0:1,samples=2,style=semithick,color={rgb,255:red,28; green,255; blue,0}]({774.9455 + 15.3997*x},{529.7989 + -8.0934*x});  
\end{axis} 
\end{tikzpicture} 

  \noindent} \resizebox {!}{0.18\textwidth} {\begin{tikzpicture} 
\begin{axis}[y dir=reverse, 
 xmin=1,xmax=960, 
 ymin=1,ymax=600, 
 xticklabels = \empty, yticklabels = \empty, 
 grid=none, axis equal image] 
\addplot graphics[xmin=1,xmax=960,ymin=1,ymax=600] {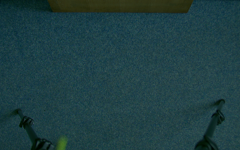}; 
\addplot [<-,>={Latex[length=1.5mm,width=0.5mm,angle'=25,open,round]},,domain=0:1,samples=10,style=semithick,color={rgb,255:red,255; green,255; blue,0}]({259.0845 + -12.3549*x + -0.91212*x^2},{491.2014 + 33.97*x + 1.8861*x^2});  
\addplot [->,>={Latex[length=1.5mm,width=0.5mm,angle'=25,open,round]},,domain=0:1,samples=2,style=semithick,color={rgb,255:red,52; green,255; blue,0}]({281.1847 + 20.923*x},{454.6937 + -73.9462*x});  
\addplot [->,>={Latex[length=1.5mm,width=0.5mm,angle'=25,open,round]},,domain=0:1,samples=2,style=semithick,color={rgb,255:red,45; green,255; blue,0}]({302.447 + 22.8387*x},{379.7659 + -81.3261*x});  
\addplot [->,>={Latex[length=1.5mm,width=0.5mm,angle'=25,open,round]},,domain=0:1,samples=2,style=semithick,color={rgb,255:red,41; green,255; blue,0}]({325.4811 + 20.9677*x},{295.9038 + -76.9177*x});  
\addplot [->,>={Latex[length=1.5mm,width=0.5mm,angle'=25,open,round]},,domain=0:1,samples=2,style=semithick,color={rgb,255:red,53; green,255; blue,0}]({346.9142 + 17.803*x},{216.4682 + -72.782*x});  
\addplot [->,>={Latex[length=1.5mm,width=0.5mm,angle'=25,open,round]},,domain=0:1,samples=2,style=semithick,color={rgb,255:red,65; green,255; blue,0}]({364.6658 + 16.7426*x},{141.4004 + -69.6627*x});  
\addplot [<-,>={Latex[length=1.5mm,width=0.5mm,angle'=25,open,round]},,domain=0:1,samples=2,style=semithick,color={rgb,255:red,32; green,255; blue,0}]({399.8167 + -8.2243*x},{115.7638 + -21.2995*x});  
\addplot [<-,>={Latex[length=1.5mm,width=0.5mm,angle'=25,open,round]},,domain=0:1,samples=2,style=semithick,color={rgb,255:red,25; green,255; blue,0}]({408.9166 + -7.7282*x},{139.7433 + -18.5304*x});  
\addplot [<-,>={Latex[length=1.5mm,width=0.5mm,angle'=25,open,round]},,domain=0:1,samples=2,style=semithick,color={rgb,255:red,29; green,255; blue,0}]({413.7685 + -4.9325*x},{160.5624 + -18.2879*x});  
\addplot [<-,>={Latex[length=1.5mm,width=0.5mm,angle'=25,open,round]},,domain=0:1,samples=2,style=semithick,color={rgb,255:red,14; green,255; blue,0}]({417.3869 + -4.158*x},{181.411 + -18.0786*x});  
\addplot [<-,>={Latex[length=1.5mm,width=0.5mm,angle'=25,open,round]},,domain=0:1,samples=2,style=semithick,color={rgb,255:red,34; green,255; blue,0}]({420.0884 + -3.4356*x},{200.983 + -17.9165*x});  
\addplot [<-,>={Latex[length=1.5mm,width=0.5mm,angle'=25,open,round]},,domain=0:1,samples=2,style=semithick,color={rgb,255:red,10; green,255; blue,0}]({422.1531 + -2.5164*x},{220.6286 + -16.5734*x});  
\addplot [<-,>={Latex[length=1.5mm,width=0.5mm,angle'=25,open,round]},,domain=0:1,samples=2,style=semithick,color={rgb,255:red,30; green,255; blue,0}]({425.4263 + -3.1455*x},{238.6181 + -15.2789*x});  
\addplot [<-,>={Latex[length=1.5mm,width=0.5mm,angle'=25,open,round]},,domain=0:1,samples=2,style=semithick,color={rgb,255:red,22; green,255; blue,0}]({429.8571 + -4.6342*x},{253.6992 + -14.9584*x});  
\end{axis} 
\end{tikzpicture} 

  \noindent} \resizebox {!}{0.18\textwidth} {\begin{tikzpicture} 
\begin{axis}[y dir=reverse, 
 xmin=1,xmax=1920, 
 ymin=1,ymax=1080, 
 xticklabels = \empty, yticklabels = \empty, 
 grid=none, axis equal image] 
\addplot graphics[xmin=1,xmax=1920,ymin=1,ymax=1080] {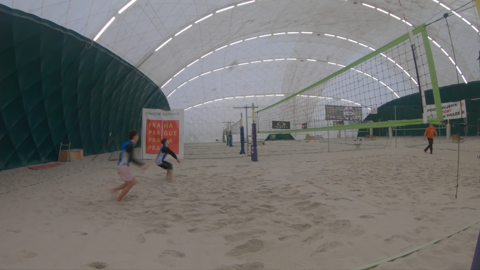}; 
\addplot [,domain=0:1,samples=10,style=semithick,color={rgb,255:red,164; green,255; blue,0}]({877.1215 + -20.6596*x + 5.6548*x^2},{327.2699 + 23.9206*x + -5.0955*x^2});  
\addplot [,domain=0:1,samples=10,style=semithick,color={rgb,255:red,121; green,255; blue,0}]({850.2903 + -19.1259*x + 1.6565*x^2},{361.623 + 27.7826*x + -3.4516*x^2});  
\addplot [->,>={Latex[length=1.5mm,width=0.5mm,angle'=25,open,round]},,domain=0:1,samples=10,style=semithick,color={rgb,255:red,124; green,255; blue,0}]({819.8589 + -16.4559*x + -0.77107*x^2},{404.8407 + 30.7786*x + -0.83269*x^2});  
\addplot [<-,>={Latex[length=1.3722mm,width=0.5mm,angle'=25,open,round]},,domain=0:1,samples=2,style=semithick,color={rgb,255:red,113; green,255; blue,0}]({774.9845 + 21.3563*x},{480.5974 + -35.194*x});  
\addplot [<-,>={Latex[length=1.4751mm,width=0.5mm,angle'=25,open,round]},,domain=0:1,samples=2,style=semithick,color={rgb,255:red,211; green,255; blue,0}]({747.66 + 22.6332*x},{526 + -38.0279*x});  
\addplot [<-,>={Latex[length=1.5mm,width=0.5mm,angle'=25,open,round]},,domain=0:1,samples=2,style=semithick,color={rgb,255:red,244; green,255; blue,0}]({719.1332 + 23.6591*x},{573.9848 + -39.8111*x});  
\addplot [<-,>={Latex[length=1.5mm,width=0.5mm,angle'=25,open,round]},,domain=0:1,samples=2,style=semithick,color=red]({689.0683 + 24.5608*x},{624.959 + -42.3034*x});  
\addplot [<-,>={Latex[length=1.3mm,width=0.5mm,angle'=25,open,round]},,domain=0:1,samples=10,style=semithick,color={rgb,255:red,72; green,255; blue,0}]({751.9882 + -5.0159*x + 0.34675*x^2},{374.5312 + 26.0433*x + 2.6592*x^2});  
\addplot [->,>={Latex[length=1.118mm,width=0.5mm,angle'=25,open,round]},,domain=0:1,samples=2,style=semithick,color={rgb,255:red,20; green,255; blue,0}]({753.5248 + 5.9004*x},{366.0938 + -33.0178*x});  
\addplot [->,>={Latex[length=1.0107mm,width=0.5mm,angle'=25,open,round]},,domain=0:1,samples=2,style=semithick,color={rgb,255:red,43; green,255; blue,0}]({759.9001 + 5.3908*x},{328.3916 + -29.8391*x});  
\addplot [->,>={Latex[length=0.87251mm,width=0.5mm,angle'=25,open,round]},,domain=0:1,samples=2,style=semithick,color={rgb,255:red,79; green,255; blue,0}]({766.5554 + 4.9706*x},{294.0107 + -25.6991*x});  
\addplot [->,>={Latex[length=0.62603mm,width=0.5mm,angle'=25,open,round]},,domain=0:1,samples=2,style=semithick,color={rgb,255:red,85; green,255; blue,0}]({778.2224 + 2.2926*x},{259.8348 + -18.6405*x});  
\addplot [->,>={Latex[length=0.59762mm,width=0.5mm,angle'=25,open,round]},,domain=0:1,samples=2,style=semithick,color={rgb,255:red,65; green,255; blue,0}]({782.8768 + 4.9441*x},{232.895 + -17.2333*x});  
\addplot [->,>={Latex[length=0.58421mm,width=0.5mm,angle'=25,open,round]},,domain=0:1,samples=2,style=semithick,color={rgb,255:red,50; green,255; blue,0}]({789.7628 + 5.0053*x},{208.7273 + -16.7964*x});  
\addplot [->,>={Latex[length=0.51071mm,width=0.5mm,angle'=25,open,round]},,domain=0:1,samples=2,style=semithick,color={rgb,255:red,35; green,255; blue,0}]({795.858 + 5.6996*x},{186.3439 + -14.2215*x});  
\addplot [->,>={Latex[length=0.5mm,width=0.5mm,angle'=25,open,round]},,domain=0:1,samples=2,style=semithick,color={rgb,255:red,45; green,255; blue,0}]({803.1601 + 6.4528*x},{166.8397 + -12.5389*x});  
\addplot [->,>={Latex[length=0.5mm,width=0.5mm,angle'=25,open,round]},,domain=0:1,samples=2,style=semithick,color={rgb,255:red,44; green,255; blue,0}]({811.676 + 4.6996*x},{150.093 + -8.8945*x});  
\addplot [->,>={Latex[length=0.5mm,width=0.5mm,angle'=25,open,round]},,domain=0:1,samples=2,style=semithick,color={rgb,255:red,40; green,255; blue,0}]({818.8276 + 4.6136*x},{137.8929 + -7.4294*x});  
\addplot [->,>={Latex[length=0.5mm,width=0.5mm,angle'=25,open,round]},,domain=0:1,samples=2,style=semithick,color={rgb,255:red,71; green,255; blue,0}]({826.5 + 6.0002*x},{122.8883 + -0.0008295*x});  
\addplot [->,>={Latex[length=0.5mm,width=0.5mm,angle'=25,open,round]},,domain=0:1,samples=2,style=semithick,color={rgb,255:red,65; green,255; blue,0}]({833.9095 + 4.6623*x},{116.9836 + -0.2803*x});  
\addplot [->,>={Latex[length=0.5mm,width=0.5mm,angle'=25,open,round]},,domain=0:1,samples=2,style=semithick,color={rgb,255:red,31; green,255; blue,0}]({839.9336 + 4.6364*x},{111.1506 + -0.22817*x});  
\addplot [->,>={Latex[length=0.5mm,width=0.5mm,angle'=25,open,round]},,domain=0:1,samples=2,style=semithick,color={rgb,255:red,45; green,255; blue,0}]({847.3655 + 4.9495*x},{110.8308 + -3.9345*x});  
\addplot [<-,>={Latex[length=0.5mm,width=0.5mm,angle'=25,open,round]},,domain=0:1,samples=2,style=semithick,color={rgb,255:red,74; green,255; blue,0}]({858.3366 + -3.6606*x},{112.2786 + -5.5655*x});  
\addplot [->,>={Latex[length=0.5mm,width=0.5mm,angle'=25,open,round]},,domain=0:1,samples=2,style=semithick,color={rgb,255:red,34; green,255; blue,0}]({861.3703 + 5.4718*x},{110.6005 + 5.0698*x});  
\addplot [<-,>={Latex[length=0.5mm,width=0.5mm,angle'=25,open,round]},,domain=0:1,samples=2,style=semithick,color={rgb,255:red,18; green,255; blue,0}]({876.2207 + -6.4075*x},{122.8017 + -7.134*x});  
\addplot [<-,>={Latex[length=0.5mm,width=0.5mm,angle'=25,open,round]},,domain=0:1,samples=2,style=semithick,color={rgb,255:red,58; green,255; blue,0}]({885.2287 + -5.6326*x},{131.8433 + -8.2241*x});  
\addplot [<-,>={Latex[length=0.5mm,width=0.5mm,angle'=25,open,round]},,domain=0:1,samples=2,style=semithick,color={rgb,255:red,52; green,255; blue,0}]({892.4407 + -6.0465*x},{145.5299 + -11.9765*x});  
\addplot [<-,>={Latex[length=0.5mm,width=0.5mm,angle'=25,open,round]},,domain=0:1,samples=2,style=semithick,color={rgb,255:red,40; green,255; blue,0}]({899.7656 + -6.5889*x},{160.6169 + -13.2062*x});  
\addplot [<-,>={Latex[length=0.58518mm,width=0.5mm,angle'=25,open,round]},,domain=0:1,samples=2,style=semithick,color={rgb,255:red,59; green,255; blue,0}]({909.1964 + -6.232*x},{179.9254 + -16.4119*x});  
\addplot [<-,>={Latex[length=0.7431mm,width=0.5mm,angle'=25,open,round]},,domain=0:1,samples=2,style=semithick,color={rgb,255:red,21; green,255; blue,0}]({917.5587 + -7.0036*x},{204.146 + -21.1643*x});  
\addplot [<-,>={Latex[length=0.79038mm,width=0.5mm,angle'=25,open,round]},,domain=0:1,samples=2,style=semithick,color={rgb,255:red,31; green,255; blue,0}]({926.3533 + -7.0022*x},{229.6999 + -22.6538*x});  
\addplot [<-,>={Latex[length=0.93356mm,width=0.5mm,angle'=25,open,round]},,domain=0:1,samples=2,style=semithick,color={rgb,255:red,41; green,255; blue,0}]({937.138 + -8.3866*x},{260.6428 + -26.7217*x});  
\addplot [<-,>={Latex[length=1.1161mm,width=0.5mm,angle'=25,open,round]},,domain=0:1,samples=2,style=semithick,color={rgb,255:red,32; green,255; blue,0}]({945.932 + -7.8949*x},{295.2739 + -32.5402*x});  
\addplot [<-,>={Latex[length=1.1768mm,width=0.5mm,angle'=25,open,round]},,domain=0:1,samples=2,style=semithick,color={rgb,255:red,38; green,255; blue,0}]({955.3143 + -7.8645*x},{333.0424 + -34.4167*x});  
\addplot [<-,>={Latex[length=1.2747mm,width=0.5mm,angle'=25,open,round]},,domain=0:1,samples=2,style=semithick,color={rgb,255:red,36; green,255; blue,0}]({963.8573 + -7.2429*x},{375.124 + -37.5496*x});  
\addplot [<-,>={Latex[length=1.3238mm,width=0.5mm,angle'=25,open,round]},,domain=0:1,samples=2,style=semithick,color={rgb,255:red,40; green,255; blue,0}]({973.2991 + -7.4603*x},{420.0384 + -39.0076*x});  
\addplot [<-,>={Latex[length=1.4219mm,width=0.5mm,angle'=25,open,round]},,domain=0:1,samples=2,style=semithick,color={rgb,255:red,72; green,255; blue,0}]({983.149 + -8.253*x},{470.872 + -41.8515*x});  
\end{axis} 
\end{tikzpicture} 

  \noindent} \\
 & \resizebox {!}{0.19\textwidth} {\begin{tikzpicture} 
\begin{axis}[y dir=reverse, 
 xmin=1,xmax=960, 
 ymin=1,ymax=600, 
 xticklabels = \empty, yticklabels = \empty, 
 grid=none, axis equal image] 
\addplot graphics[xmin=1,xmax=960,ymin=1,ymax=600] {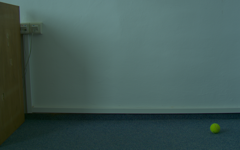}; 
\addplot [<-,>={Latex[length=1.5mm,width=0.5mm,angle'=25,open,round]},,domain=0:1,samples=2,style=semithick,color={rgb,255:red,144; green,255; blue,0}]({347.7874 + 45.8641*x},{526.6843 + 1.6346*x});  
\addplot [>={Latex[length=1.5mm,width=0.5mm,angle'=25,open,round]},,domain=0:1,samples=2,style=semithick,color={rgb,255:red,144; green,255; blue,0}]({393.6516 + 92.7349*x},{528.3189 + -14.3246*x});  
\addplot [<-,>={Latex[length=1.5mm,width=0.5mm,angle'=25,open,round]},,domain=0:1,samples=10,style=semithick,color={rgb,255:red,121; green,255; blue,0}]({214.5014 + 143.7905*x + -20.7155*x^2},{506.9972 + 6.831*x + 6.2757*x^2});  
\addplot [<-,>={Latex[length=1.5mm,width=0.5mm,angle'=25,open,round]},,domain=0:1,samples=2,style=semithick,color={rgb,255:red,75; green,255; blue,0}]({94.4598 + 117.0571*x},{510.4598 + -4.5205*x});  
\addplot [->,>={Latex[length=1.5mm,width=0.5mm,angle'=25,open,round]},,domain=0:1,samples=10,style=semithick,color={rgb,255:red,157; green,255; blue,0}]({54.0446 + 65.1362*x + -5.0627*x^2},{517.7723 + -11.4484*x + 7.1573*x^2});  
\addplot [->,>={Latex[length=1.5mm,width=0.5mm,angle'=25,open,round]},,domain=0:1,samples=10,style=semithick,color={rgb,255:red,77; green,255; blue,0}]({117.0541 + 100.014*x + -14.2962*x^2},{513.339 + 3.6983*x + 6.8758*x^2});  
\addplot [->,>={Latex[length=1.5mm,width=0.5mm,angle'=25,open,round]},,domain=0:1,samples=10,style=semithick,color={rgb,255:red,96; green,255; blue,0}]({204.2362 + 82.2306*x + -19.4537*x^2},{527.3549 + -23.2746*x + 12.2812*x^2});  
\addplot [->,>={Latex[length=1.5mm,width=0.5mm,angle'=25,open,round]},,domain=0:1,samples=2,style=semithick,color={rgb,255:red,59; green,255; blue,0}]({267.1342 + 55.3651*x},{515.1147 + 8.3901*x});  
\addplot [->,>={Latex[length=1.5mm,width=0.5mm,angle'=25,open,round]},,domain=0:1,samples=10,style=semithick,color={rgb,255:red,88; green,255; blue,0}]({323.8054 + 55.7674*x + 1.4774*x^2},{522.4554 + -11.0464*x + 8.487*x^2});  
\addplot [->,>={Latex[length=1.5mm,width=0.5mm,angle'=25,open,round]},,domain=0:1,samples=2,style=semithick,color={rgb,255:red,73; green,255; blue,0}]({378.0065 + 55.5814*x},{522.0709 + -5.1155*x});  
\addplot [->,>={Latex[length=1.5mm,width=0.5mm,angle'=25,open,round]},,domain=0:1,samples=10,style=semithick,color={rgb,255:red,71; green,255; blue,0}]({433.5333 + 48.3787*x + 4.0969*x^2},{517.3069 + 8.505*x + -9.7617*x^2});  
\addplot [->,>={Latex[length=1.5mm,width=0.5mm,angle'=25,open,round]},,domain=0:1,samples=10,style=semithick,color={rgb,255:red,70; green,255; blue,0}]({486.0066 + 50.7962*x + 1.5523*x^2},{517.419 + 2.2744*x + -6.0268*x^2});  
\addplot [->,>={Latex[length=1.5mm,width=0.5mm,angle'=25,open,round]},,domain=0:1,samples=10,style=semithick,color={rgb,255:red,70; green,255; blue,0}]({538.4849 + 48.6092*x + 2.3418*x^2},{514.7067 + 3.8638*x + -7.3825*x^2});  
\addplot [->,>={Latex[length=1.5mm,width=0.5mm,angle'=25,open,round]},,domain=0:1,samples=10,style=semithick,color={rgb,255:red,73; green,255; blue,0}]({589.4484 + 48.868*x + 0.59615*x^2},{512.4062 + 3.1287*x + -6.042*x^2});  
\addplot [->,>={Latex[length=1.5mm,width=0.5mm,angle'=25,open,round]},,domain=0:1,samples=2,style=semithick,color={rgb,255:red,72; green,255; blue,0}]({638.9574 + 49.5656*x},{510.6652 + -0.90446*x});  
\addplot [->,>={Latex[length=1.5mm,width=0.5mm,angle'=25,open,round]},,domain=0:1,samples=2,style=semithick,color={rgb,255:red,74; green,255; blue,0}]({686.8981 + 48.1831*x},{509.5964 + -2.5781*x});  
\addplot [->,>={Latex[length=1.5mm,width=0.5mm,angle'=25,open,round]},,domain=0:1,samples=2,style=semithick,color={rgb,255:red,76; green,255; blue,0}]({734.8422 + 46.6882*x},{507.8714 + -3.4613*x});  
\addplot [->,>={Latex[length=1.5mm,width=0.5mm,angle'=25,open,round]},,domain=0:1,samples=2,style=semithick,color={rgb,255:red,69; green,255; blue,0}]({781.4153 + 45.1098*x},{505.5194 + -2.5819*x});  
\addplot [->,>={Latex[length=1.5mm,width=0.5mm,angle'=25,open,round]},,domain=0:1,samples=2,style=semithick,color={rgb,255:red,75; green,255; blue,0}]({826.3967 + 45.1766*x},{503.1529 + -1.9882*x});  
\addplot [->,>={Latex[length=1.4115mm,width=0.5mm,angle'=25,open,round]},,domain=0:1,samples=2,style=semithick,color={rgb,255:red,78; green,255; blue,0}]({871.3735 + 42.2096*x},{500.9255 + -3.3914*x});  
\addplot [->,>={Latex[length=1.2025mm,width=0.5mm,angle'=25,open,round]},,domain=0:1,samples=2,style=semithick,color={rgb,255:red,68; green,255; blue,0}]({913.4759 + 36.0584*x},{497.2255 + -1.1224*x});  
\end{axis} 
\end{tikzpicture} 

  \noindent}   \resizebox {!}{0.19\textwidth} {\begin{tikzpicture} 
\begin{axis}[y dir=reverse, 
 xmin=1,xmax=1280, 
 ymin=1,ymax=960, 
 xticklabels = \empty, yticklabels = \empty, 
 grid=none, axis equal image] 
\addplot graphics[xmin=1,xmax=1280,ymin=1,ymax=960] {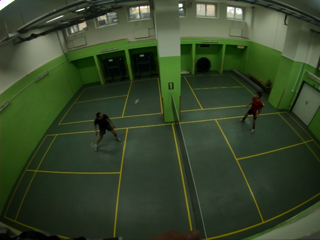}; 
\addplot [<-,>={Latex[length=1.5mm,width=0.5mm,angle'=25,open,round]},,domain=0:1,samples=2,style=semithick,color={rgb,255:red,75; green,255; blue,0}]({930.9835 + 18.1008*x},{286.8701 + 4.0493*x});  
\addplot [>={Latex[length=1.5mm,width=0.5mm,angle'=25,open,round]},,domain=0:1,samples=2,style=semithick,color={rgb,255:red,75; green,255; blue,0}]({949.0844 + 18.0297*x},{290.9194 + 4.0972*x});  
\addplot [<-,>={Latex[length=1.5mm,width=0.5mm,angle'=25,open,round]},,domain=0:1,samples=2,style=semithick,color={rgb,255:red,79; green,255; blue,0}]({884.9329 + 43.3373*x},{278.3535 + 8.222*x});  
\addplot [<-,>={Latex[length=1.5mm,width=0.5mm,angle'=25,open,round]},,domain=0:1,samples=10,style=semithick,color={rgb,255:red,112; green,255; blue,0}]({839.002 + 38.7522*x + 5.3868*x^2},{272.9967 + 1.8245*x + 3.3711*x^2});  
\addplot [<-,>={Latex[length=1.5mm,width=0.5mm,angle'=25,open,round]},,domain=0:1,samples=2,style=semithick,color={rgb,255:red,114; green,255; blue,0}]({793.0837 + 44.9089*x},{269.8461 + 3.2563*x});  
\addplot [>={Latex[length=1.5mm,width=0.5mm,angle'=25,open,round]},,domain=0:1,samples=2,style=semithick,color={rgb,255:red,48; green,255; blue,0}]({792.417 + -23.6489*x},{268.6717 + -0.53124*x});  
\addplot [->,>={Latex[length=1.5mm,width=0.5mm,angle'=25,open,round]},,domain=0:1,samples=2,style=semithick,color={rgb,255:red,48; green,255; blue,0}]({768.7681 + -17.4004*x},{268.1405 + -0.1509*x});  
\addplot [<-,>={Latex[length=1.5mm,width=0.5mm,angle'=25,open,round]},,domain=0:1,samples=2,style=semithick,color={rgb,255:red,77; green,255; blue,0}]({706.9558 + 43.1435*x},{271.3055 + -2.745*x});  
\addplot [<-,>={Latex[length=1.5mm,width=0.5mm,angle'=25,open,round]},,domain=0:1,samples=2,style=semithick,color={rgb,255:red,76; green,255; blue,0}]({663.9855 + 42.1896*x},{274.8475 + -4.0006*x});  
\addplot [<-,>={Latex[length=1.5mm,width=0.5mm,angle'=25,open,round]},,domain=0:1,samples=2,style=semithick,color={rgb,255:red,46; green,255; blue,0}]({628.0936 + 20.1896*x},{281.7996 + -3.8693*x});  
\addplot [>={Latex[length=1.5mm,width=0.5mm,angle'=25,open,round]},,domain=0:1,samples=2,style=semithick,color={rgb,255:red,46; green,255; blue,0}]({648.2832 + 16.5317*x},{277.9303 + -3.1433*x});  
\addplot [<-,>={Latex[length=1.5mm,width=0.5mm,angle'=25,open,round]},,domain=0:1,samples=10,style=semithick,color={rgb,255:red,90; green,255; blue,0}]({587.7382 + 37.3922*x + 1.8139*x^2},{291.1791 + -12.1157*x + 3.4948*x^2});  
\addplot [<-,>={Latex[length=1.5mm,width=0.5mm,angle'=25,open,round]},,domain=0:1,samples=2,style=semithick,color={rgb,255:red,76; green,255; blue,0}]({554.0733 + 35.2845*x},{300.2588 + -9.9958*x});  
\addplot [<-,>={Latex[length=1.5mm,width=0.5mm,angle'=25,open,round]},,domain=0:1,samples=2,style=semithick,color={rgb,255:red,97; green,255; blue,0}]({522.544 + 32.3678*x},{311.647 + -10.9088*x});  
\addplot [<-,>={Latex[length=1.5mm,width=0.5mm,angle'=25,open,round]},,domain=0:1,samples=10,style=semithick,color={rgb,255:red,123; green,255; blue,0}]({491.9723 + 24.8323*x + 5.2865*x^2},{324.9464 + -13.4146*x + 0.73698*x^2});  
\addplot [<-,>={Latex[length=1.5mm,width=0.5mm,angle'=25,open,round]},,domain=0:1,samples=2,style=semithick,color={rgb,255:red,62; green,255; blue,0}]({465.9994 + 26.4246*x},{337.9988 + -14.2103*x});  
\addplot [<-,>={Latex[length=1.5mm,width=0.5mm,angle'=25,open,round]},,domain=0:1,samples=2,style=semithick,color={rgb,255:red,71; green,255; blue,0}]({440.8424 + 26.1194*x},{352.7394 + -15.8027*x});  
\addplot [<-,>={Latex[length=1.4997mm,width=0.5mm,angle'=25,open,round]},,domain=0:1,samples=2,style=semithick,color={rgb,255:red,103; green,255; blue,0}]({417.4976 + 24.3558*x},{368.3009 + -17.5048*x});  
\addplot [<-,>={Latex[length=1.3098mm,width=0.5mm,angle'=25,open,round]},,domain=0:1,samples=10,style=semithick,color={rgb,255:red,108; green,255; blue,0}]({399.4104 + 17.3159*x + 2.705*x^2},{383.4433 + -18.6363*x + 1.8413*x^2});  
\addplot [<-,>={Latex[length=1.2377mm,width=0.5mm,angle'=25,open,round]},,domain=0:1,samples=2,style=semithick,color={rgb,255:red,53; green,255; blue,0}]({383.2386 + 17.3631*x},{398.2506 + -17.6426*x});  
\addplot [<-,>={Latex[length=1.1391mm,width=0.5mm,angle'=25,open,round]},,domain=0:1,samples=2,style=semithick,color={rgb,255:red,70; green,255; blue,0}]({368.1933 + 14.7394*x},{414.3156 + -17.3727*x});  
\addplot [<-,>={Latex[length=1.1099mm,width=0.5mm,angle'=25,open,round]},,domain=0:1,samples=2,style=semithick,color={rgb,255:red,76; green,255; blue,0}]({355.4332 + 13.4042*x},{430.5706 + -17.6938*x});  
\addplot [<-,>={Latex[length=0.98467mm,width=0.5mm,angle'=25,open,round]},,domain=0:1,samples=2,style=semithick,color={rgb,255:red,121; green,255; blue,0}]({344.108 + 10.8995*x},{446.4073 + -16.4023*x});  
\addplot [,domain=0:1,samples=10,style=semithick,color={rgb,255:red,80; green,255; blue,0}]({344.6511 + -5.6498*x + -0.33557*x^2},{447.2389 + 9.6438*x + -1.1623*x^2});  
\addplot [<-,>={Latex[length=1.5mm,width=0.5mm,angle'=25,open,round]},,domain=0:1,samples=2,style=semithick,color={rgb,255:red,92; green,255; blue,0}]({449.5292 + -22.1272*x},{389.2176 + 19.2721*x});  
\addplot [>={Latex[length=1.5mm,width=0.5mm,angle'=25,open,round]},,domain=0:1,samples=2,style=semithick,color={rgb,255:red,92; green,255; blue,0}]({427.402 + -17.992*x},{408.4897 + 13.1648*x});  
\addplot [<-,>={Latex[length=1.5mm,width=0.5mm,angle'=25,open,round]},,domain=0:1,samples=10,style=semithick,color={rgb,255:red,165; green,255; blue,0}]({509.8024 + -71.738*x + 15.781*x^2},{351.5975 + 35.0969*x + 0.12803*x^2});  
\addplot [->,>={Latex[length=1.5mm,width=0.5mm,angle'=25,open,round]},,domain=0:1,samples=2,style=semithick,color={rgb,255:red,100; green,255; blue,0}]({511.9874 + 55.2212*x},{351.9728 + -25.5214*x});  
\addplot [->,>={Latex[length=1.5mm,width=0.5mm,angle'=25,open,round]},,domain=0:1,samples=2,style=semithick,color={rgb,255:red,84; green,255; blue,0}]({566.4359 + 52.5428*x},{326.0837 + -21.1365*x});  
\addplot [<-,>={Latex[length=1.5mm,width=0.5mm,angle'=25,open,round]},,domain=0:1,samples=10,style=semithick,color={rgb,255:red,87; green,255; blue,0}]({662.7751 + -40.5491*x + -4.6347*x^2},{286.5384 + 20.5868*x + -2.4218*x^2});  
\addplot [->,>={Latex[length=1.5mm,width=0.5mm,angle'=25,open,round]},,domain=0:1,samples=2,style=semithick,color={rgb,255:red,102; green,255; blue,0}]({665.9969 + 40.9681*x},{285.9916 + -15.0867*x});  
\addplot [<-,>={Latex[length=1.5mm,width=0.5mm,angle'=25,open,round]},,domain=0:1,samples=10,style=semithick,color={rgb,255:red,78; green,255; blue,0}]({751.0795 + -42.6731*x + 1.4507*x^2},{259.3287 + 10.6068*x + 0.59382*x^2});  
\addplot [>={Latex[length=1.5mm,width=0.5mm,angle'=25,open,round]},,domain=0:1,samples=2,style=semithick,color={rgb,255:red,58; green,255; blue,0}]({750.3599 + 18.5144*x},{259.312 + -4.7974*x});  
\addplot [->,>={Latex[length=1.5mm,width=0.5mm,angle'=25,open,round]},,domain=0:1,samples=2,style=semithick,color={rgb,255:red,58; green,255; blue,0}]({768.8742 + 16.0086*x},{254.5146 + -3.6978*x});  
\addplot [->,>={Latex[length=1.5mm,width=0.5mm,angle'=25,open,round]},,domain=0:1,samples=2,style=semithick,color={rgb,255:red,101; green,255; blue,0}]({786.7613 + 33.381*x},{251.6292 + -5.8117*x});  
\addplot [->,>={Latex[length=1.5mm,width=0.5mm,angle'=25,open,round]},,domain=0:1,samples=2,style=semithick,color={rgb,255:red,100; green,255; blue,0}]({819.9278 + 29.9959*x},{246.2879 + -3.0404*x});  
\addplot [->,>={Latex[length=1.404mm,width=0.5mm,angle'=25,open,round]},,domain=0:1,samples=10,style=semithick,color={rgb,255:red,58; green,255; blue,0}]({848.3212 + 26.8582*x + 0.85455*x^2},{243.2443 + -7.2039*x + 7.8239*x^2});  
\addplot [->,>={Latex[length=1.2602mm,width=0.5mm,angle'=25,open,round]},,domain=0:1,samples=2,style=semithick,color={rgb,255:red,83; green,255; blue,0}]({875.9446 + 25.1525*x},{241.8731 + 1.5951*x});  
\addplot [->,>={Latex[length=1.2926mm,width=0.5mm,angle'=25,open,round]},,domain=0:1,samples=10,style=semithick,color={rgb,255:red,74; green,255; blue,0}]({897.8117 + 23.7648*x + 1.1397*x^2},{243.7236 + -4.2555*x + 9.0615*x^2});  
\addplot [->,>={Latex[length=1.009mm,width=0.5mm,angle'=25,open,round]},,domain=0:1,samples=2,style=semithick,color={rgb,255:red,51; green,255; blue,0}]({921.7828 + 19.4461*x},{245.7835 + 5.391*x});  
\addplot [->,>={Latex[length=0.98296mm,width=0.5mm,angle'=25,open,round]},,domain=0:1,samples=2,style=semithick,color={rgb,255:red,57; green,255; blue,0}]({940.904 + 18.4287*x},{250.2584 + 6.846*x});  
\end{axis} 
\end{tikzpicture} 

  \noindent} \resizebox {!}{0.19\textwidth} {\begin{tikzpicture} 
\begin{axis}[y dir=reverse, 
 xmin=1,xmax=1920, 
 ymin=1,ymax=1080, 
 xticklabels = \empty, yticklabels = \empty, 
 grid=none, axis equal image] 
\addplot graphics[xmin=1,xmax=1920,ymin=1,ymax=1080] {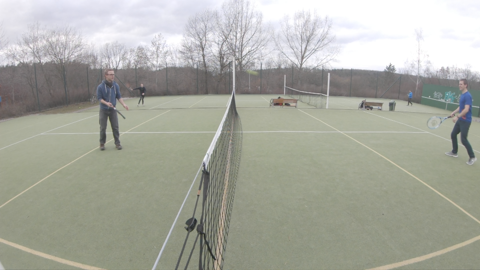}; 

\addplot [>={Latex[length=1.4mm,width=0.5mm,angle'=25,open,round]},,domain=0:1,samples=300,style=semithick,color=blue]({1569.6857 + -696.9964*x + -130.3767*x^2},{367.4297 + -513.8605*x + 768.4476*x^2});
\addplot [<-,>={Latex[length=1.4mm,width=0.5mm,angle'=25,open,round]},,domain=0:1,samples=2,style=semithick,color=blue]({626.8061 + 78.9185*x},{471.7071 + 91.1562*x});
\addplot [>={Latex[length=1.4mm,width=0.5mm,angle'=25,open,round]},,domain=0:1,samples=2,style=semithick,color=blue]({705.7246 + 35.2557*x},{562.8633 + 64.1934*x}); 

\addplot [->,>={Latex[length=0.95mm,width=0.5mm,angle'=25,open,round]},,domain=0:1,samples=10,style=semithick,color={rgb,255:red,118; green,255; blue,0}]({1568.1737 + -13.3418*x + 0.80705*x^2},{364.2985 + -9.9037*x + 0.46311*x^2});  
\addplot [<-,>={Latex[length=1.1925mm,width=0.5mm,angle'=25,open,round]},,domain=0:1,samples=2,style=semithick,color={rgb,255:red,44; green,255; blue,0}]({1532.086 + 19.7754*x},{338.8724 + 13.3331*x});  
\addplot [<-,>={Latex[length=1.2093mm,width=0.5mm,angle'=25,open,round]},,domain=0:1,samples=2,style=semithick,color={rgb,255:red,47; green,255; blue,0}]({1506.8293 + 21.073*x},{324.3031 + 11.8704*x});  
\addplot [<-,>={Latex[length=1.2083mm,width=0.5mm,angle'=25,open,round]},,domain=0:1,samples=2,style=semithick,color={rgb,255:red,36; green,255; blue,0}]({1480.1819 + 21.9343*x},{311.6066 + 10.1421*x});  
\addplot [<-,>={Latex[length=1.2347mm,width=0.5mm,angle'=25,open,round]},,domain=0:1,samples=2,style=semithick,color={rgb,255:red,54; green,255; blue,0}]({1452.903 + 23.1532*x},{301.2616 + 8.5869*x});  
\addplot [<-,>={Latex[length=1.25mm,width=0.5mm,angle'=25,open,round]},,domain=0:1,samples=2,style=semithick,color={rgb,255:red,49; green,255; blue,0}]({1424.02 + 24.0157*x},{292.9307 + 6.9456*x});  
\addplot [<-,>={Latex[length=1.3237mm,width=0.5mm,angle'=25,open,round]},,domain=0:1,samples=2,style=semithick,color={rgb,255:red,62; green,255; blue,0}]({1394.0761 + 25.9283*x},{286.631 + 5.3479*x});  
\addplot [<-,>={Latex[length=1.2586mm,width=0.5mm,angle'=25,open,round]},,domain=0:1,samples=2,style=semithick,color={rgb,255:red,53; green,255; blue,0}]({1363.9742 + 25.0587*x},{283.2713 + 2.382*x});  
\addplot [<-,>={Latex[length=1.1508mm,width=0.5mm,angle'=25,open,round]},,domain=0:1,samples=2,style=semithick,color={rgb,255:red,72; green,255; blue,0}]({1334.0044 + 23.0108*x},{282.8071 + 0.52437*x});  
\addplot [<-,>={Latex[length=1.2157mm,width=0.5mm,angle'=25,open,round]},,domain=0:1,samples=10,style=semithick,color={rgb,255:red,103; green,255; blue,0}]({1297.0446 + 23.721*x + 0.33611*x^2},{283.2216 + -4.2602*x + 5.6517*x^2});  
\addplot [<-,>={Latex[length=1.4134mm,width=0.5mm,angle'=25,open,round]},,domain=0:1,samples=2,style=semithick,color={rgb,255:red,42; green,255; blue,0}]({1262.9097 + 28.1046*x},{286.1617 + -3.0297*x});  
\addplot [<-,>={Latex[length=1.5mm,width=0.5mm,angle'=25,open,round]},,domain=0:1,samples=2,style=semithick,color={rgb,255:red,28; green,255; blue,0}]({1225.9591 + 31.0434*x},{292.7799 + -5.7661*x});  
\addplot [<-,>={Latex[length=1.5mm,width=0.5mm,angle'=25,open,round]},,domain=0:1,samples=2,style=semithick,color={rgb,255:red,20; green,255; blue,0}]({1188.8539 + 33.0559*x},{302.4506 + -8.7896*x});  
\addplot [<-,>={Latex[length=1.5mm,width=0.5mm,angle'=25,open,round]},,domain=0:1,samples=2,style=semithick,color={rgb,255:red,31; green,255; blue,0}]({1151.842 + 34.2316*x},{314.5212 + -11.2981*x});  
\addplot [<-,>={Latex[length=1.5mm,width=0.5mm,angle'=25,open,round]},,domain=0:1,samples=2,style=semithick,color={rgb,255:red,40; green,255; blue,0}]({1113.0765 + 36.449*x},{330.182 + -15.3116*x});  
\addplot [<-,>={Latex[length=1.5mm,width=0.5mm,angle'=25,open,round]},,domain=0:1,samples=2,style=semithick,color={rgb,255:red,32; green,255; blue,0}]({1074.5893 + 35.5942*x},{348.1264 + -16.7362*x});  
\addplot [<-,>={Latex[length=1.5mm,width=0.5mm,angle'=25,open,round]},,domain=0:1,samples=2,style=semithick,color={rgb,255:red,60; green,255; blue,0}]({1034.9666 + 36.8476*x},{369.9392 + -20.2769*x});  
\addplot [<-,>={Latex[length=1.2966mm,width=0.5mm,angle'=25,open,round]},,domain=0:1,samples=2,style=semithick,color={rgb,255:red,239; green,255; blue,0}]({1010.2408 + 22.6463*x},{384.4317 + -12.634*x});  
\addplot [->,>={Latex[length=1.5mm,width=0.5mm,angle'=25,open,round]},,domain=0:1,samples=10,style=semithick,color={rgb,255:red,117; green,255; blue,0}]({989.9166 + -30.9272*x + 3.8885*x^2},{398.9786 + 21.6364*x + -2.5676*x^2});  
\addplot [<-,>={Latex[length=1.5mm,width=0.5mm,angle'=25,open,round]},,domain=0:1,samples=2,style=semithick,color={rgb,255:red,210; green,255; blue,0}]({918.5203 + 25.4853*x},{450.3686 + -19.3612*x});  
\addplot [->,>={Latex[length=1.5mm,width=0.5mm,angle'=25,open,round]},,domain=0:1,samples=2,style=semithick,color={rgb,255:red,255; green,33; blue,0}]({917.3961 + -3.7269*x},{483.0564 + -39.9322*x});  
\addplot [<-,>={Latex[length=1.1234mm,width=0.5mm,angle'=25,open,round]},,domain=0:1,samples=10,style=semithick,color=red]({911.822 + 0.35877*x + 3.853*x^2},{456.9882 + -22.4182*x + 0.46058*x^2});  
\addplot [->,>={Latex[length=1.1234mm,width=0.5mm,angle'=25,open,round]},domain=0:1,samples=10,style=semithick,color={rgb,255:red,84; green,255; blue,0}]({843.6752 + -27.6249*x + 2.995*x^2},{516.4749 + 27.2734*x + -2.9037*x^2});  
\addplot [->,>={Latex[length=1.1234mm,width=0.5mm,angle'=25,open,round]},domain=0:1,samples=10,style=semithick,color={rgb,255:red,98; green,255; blue,0}]({808.4995 + -23.2679*x + -0.20758*x^2},{551.9792 + 24.9235*x + 0.26951*x^2});  
\addplot [->,>={Latex[length=1.1234mm,width=0.5mm,angle'=25,open,round]},domain=0:1,samples=10,style=semithick,color={rgb,255:red,61; green,255; blue,0}]({777.8273 + -27.3327*x + 2.5235*x^2},{585.3875 + 32.3328*x + -2.9076*x^2});  
\addplot [->,>={Latex[length=1.1234mm,width=0.5mm,angle'=25,open,round]},domain=0:1,samples=10,style=semithick,color={rgb,255:red,214; green,255; blue,0}]({744.174 + -4.5508*x + -0.50777*x^2},{630.8336 + -9.1226*x + -0.06154*x^2});  
\addplot [->,>={Latex[length=1.1234mm,width=0.5mm,angle'=25,open,round]},domain=0:1,samples=10,style=semithick,color={rgb,255:red,90; green,255; blue,0}]({728.0073 + -11.8982*x + 0.99787*x^2},{601.8463 + -20.0667*x + 1.9309*x^2});  
\addplot [->,>={Latex[length=1.2mm,width=0.5mm,angle'=25,open,round]},,domain=0:1,samples=10,style=semithick,color={rgb,255:red,106; green,255; blue,0}]({710.8682 + -9.0571*x + -1.5784*x^2},{575.3 + -12.3907*x + -2.8922*x^2});  
\addplot [->,>={Latex[length=1.1234mm,width=0.5mm,angle'=25,open,round]},domain=0:1,samples=10,style=semithick,color={rgb,255:red,108; green,255; blue,0}]({680.0476 + -12.4719*x + 0.029451*x^2},{531.8782 + -14.9302*x + -0.16477*x^2});  
\addplot [->,>={Latex[length=1.1234mm,width=0.5mm,angle'=25,open,round]},domain=0:1,samples=10,style=semithick,color={rgb,255:red,101; green,255; blue,0}]({662.327 + -9.2314*x + -0.33633*x^2},{511.2819 + -11.0393*x + -0.33502*x^2});  
\addplot [->,>={Latex[length=1.1234mm,width=0.5mm,angle'=25,open,round]},domain=0:1,samples=10,style=semithick,color={rgb,255:red,65; green,255; blue,0}]({648.496 + -11.9122*x + 0.35146*x^2},{493.9372 + -10.6006*x + 0.56686*x^2});  
\addplot [->,>={Latex[length=1.1234mm,width=0.5mm,angle'=25,open,round]},domain=0:1,samples=10,style=semithick,color={rgb,255:red,86; green,255; blue,0}]({631.9525 + -11.8104*x + 1.9237*x^2},{479.7054 + -9.1097*x + 1.305*x^2});  
\end{axis} 
\end{tikzpicture} 

  \noindent}  \\
\end{tabular}

\caption{Trajectory recovery for selected sequences from the TbD dataset. Trajectory Intersection over Union (TIoU) with ground truth trajectories from a high-speed camera, color coded.
Arrows indicate the direction of the motion.
The trajectory over the whole sequence can be obtained by fitting a continuous piecewise quadratic curve~\eqref{eq:fitting_curve_def} on all frames jointly, as shown in the bottom right in blue.}
\label{tbl:tbd_imgs}
\end{figure*}



\begin{figure*}
\centering
\begin{tabular}{@{}c@{}c@{}c@{}}
\resizebox {0.33\textwidth}{!} {\begin{tikzpicture} 
\begin{axis}[y dir=reverse, 
 xmin=1,xmax=1280, 
 ymin=1,ymax=720, 
 xticklabels = \empty, yticklabels = \empty, 
 grid=none, axis equal image] 
\addplot graphics[xmin=1,xmax=1280,ymin=1,ymax=720] {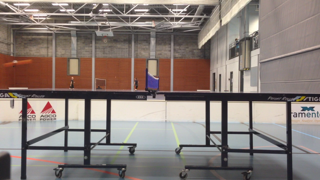}; 
\addplot [,domain=0:1,samples=2,style=semithick,color={rgb,255:red,44; green,255; blue,0}]({21.3187 + 31.3035*x},{261.056 + -5.3112*x});  
\addplot [,domain=0:1,samples=2,style=semithick,color={rgb,255:red,44; green,255; blue,0}]({52.6223 + 70.7739*x},{255.7448 + -10.6181*x});  
\addplot [,domain=0:1,samples=2,style=semithick,color={rgb,255:red,57; green,255; blue,0}]({133.1593 + 50.6346*x},{243.5295 + -6.0895*x});  
\addplot [,domain=0:1,samples=2,style=semithick,color={rgb,255:red,57; green,255; blue,0}]({183.7939 + 44.3076*x},{237.44 + -3.8745*x});  
\addplot [,domain=0:1,samples=2,style=semithick,color={rgb,255:red,70; green,255; blue,0}]({233.0892 + 64.4009*x},{233.2639 + -3.3949*x});  
\addplot [,domain=0:1,samples=2,style=semithick,color={rgb,255:red,70; green,255; blue,0}]({297.4902 + 20.8084*x},{229.8689 + -0.8858*x});  
\addplot [>={Latex[length=1.5mm,width=0.5mm,angle'=25,open,round]},,domain=0:1,samples=2,style=semithick,color={rgb,255:red,54; green,255; blue,0}]({329.86 + 46.0803*x},{228.8558 + -0.59934*x});  
\addplot [->,>={Latex[length=1.5mm,width=0.5mm,angle'=25,open,round]},,domain=0:1,samples=2,style=semithick,color={rgb,255:red,54; green,255; blue,0}]({375.9403 + 30.6745*x},{228.2565 + 1.4156*x});  
\addplot [->,>={Latex[length=1.5mm,width=0.5mm,angle'=25,open,round]},,domain=0:1,samples=2,style=semithick,color={rgb,255:red,80; green,255; blue,0}]({421.8772 + 72.1397*x},{229.7815 + 4.9726*x});  
\addplot [>={Latex[length=1.5mm,width=0.5mm,angle'=25,open,round]},,domain=0:1,samples=2,style=semithick,color={rgb,255:red,89; green,255; blue,0}]({499.9594 + 40.9913*x},{235.2985 + 4.4069*x});  
\addplot [->,>={Latex[length=1.5mm,width=0.5mm,angle'=25,open,round]},,domain=0:1,samples=2,style=semithick,color={rgb,255:red,89; green,255; blue,0}]({540.9507 + 33.1693*x},{239.7054 + 5.4871*x});  
\addplot [->,>={Latex[length=1.5mm,width=0.5mm,angle'=25,open,round]},,domain=0:1,samples=2,style=semithick,color={rgb,255:red,195; green,255; blue,0}]({624.7286 + 29.371*x},{255.2736 + 6.2587*x});  
\addplot [>={Latex[length=1.5mm,width=0.5mm,angle'=25,open,round]},,domain=0:1,samples=2,style=semithick,color={rgb,255:red,46; green,255; blue,0}]({658.1295 + 23.7038*x},{262.1474 + 5.9887*x});  
\addplot [->,>={Latex[length=1.5mm,width=0.5mm,angle'=25,open,round]},,domain=0:1,samples=2,style=semithick,color={rgb,255:red,46; green,255; blue,0}]({681.8333 + 40.3761*x},{268.136 + 11.6866*x});  
\addplot [->,>={Latex[length=1.5mm,width=0.5mm,angle'=25,open,round]},,domain=0:1,samples=2,style=semithick,color={rgb,255:red,62; green,255; blue,0}]({731.2308 + 69.3048*x},{282.3642 + 25.1604*x});  
\addplot [<-,>={Latex[length=1.5mm,width=0.5mm,angle'=25,open,round]},,domain=0:1,samples=10,style=semithick,color={rgb,255:red,169; green,255; blue,0}]({844.2525 + -39.2292*x + -2.6326*x^2},{326.4891 + -19.0406*x + 1.4833*x^2});  
\addplot [>={Latex[length=1.5mm,width=0.5mm,angle'=25,open,round]},,domain=0:1,samples=2,style=semithick,color={rgb,255:red,58; green,255; blue,0}]({923.9315 + 20.1593*x},{352.3132 + -12.385*x});  
\addplot [->,>={Latex[length=1.5mm,width=0.5mm,angle'=25,open,round]},,domain=0:1,samples=2,style=semithick,color={rgb,255:red,58; green,255; blue,0}]({944.0908 + 23.9741*x},{339.9282 + -13.2193*x});  
\addplot [<-,>={Latex[length=1.5mm,width=0.5mm,angle'=25,open,round]},,domain=0:1,samples=10,style=semithick,color={rgb,255:red,63; green,255; blue,0}]({1017.2442 + -67.3764*x + 17.5885*x^2},{302.5734 + 28.8549*x + -5.4057*x^2});  
\addplot [->,>={Latex[length=1.5mm,width=0.5mm,angle'=25,open,round]},,domain=0:1,samples=2,style=semithick,color={rgb,255:red,101; green,255; blue,0}]({1016.9343 + 46.6441*x},{301.8195 + -16.9779*x});  
\addplot [->,>={Latex[length=1.5mm,width=0.5mm,angle'=25,open,round]},,domain=0:1,samples=2,style=semithick,color={rgb,255:red,92; green,255; blue,0}]({1060.748 + 48.0089*x},{285.0671 + -12.9671*x});  
\addplot [->,>={Latex[length=1.5mm,width=0.5mm,angle'=25,open,round]},,domain=0:1,samples=2,style=semithick,color={rgb,255:red,78; green,255; blue,0}]({1106.8821 + 46.9865*x},{272.3141 + -8.0785*x});  
\addplot [->,>={Latex[length=1.5mm,width=0.5mm,angle'=25,open,round]},,domain=0:1,samples=2,style=semithick,color={rgb,255:red,100; green,255; blue,0}]({1151.9562 + 45.0348*x},{264.4456 + -3.5603*x});  
\addplot [->,>={Latex[length=1.5mm,width=0.5mm,angle'=25,open,round]},,domain=0:1,samples=2,style=semithick,color={rgb,255:red,86; green,255; blue,0}]({1195.9956 + 46.0019*x},{261.1031 + 1.9553*x});  
\addplot [->,>={Latex[length=1.5mm,width=0.5mm,angle'=25,open,round]},,domain=0:1,samples=2,style=semithick,color={rgb,255:red,56; green,255; blue,0}]({1238.2156 + 37.9416*x},{263.1795 + 4.4932*x});  
\end{axis} 
\end{tikzpicture} 

  \noindent} & \resizebox {0.33\textwidth}{!} {\begin{tikzpicture} 
\begin{axis}[y dir=reverse, 
 xmin=1,xmax=1920, 
 ymin=1,ymax=1080, 
 xticklabels = \empty, yticklabels = \empty, 
 grid=none, axis equal image] 
\addplot graphics[xmin=1,xmax=1920,ymin=1,ymax=1080] {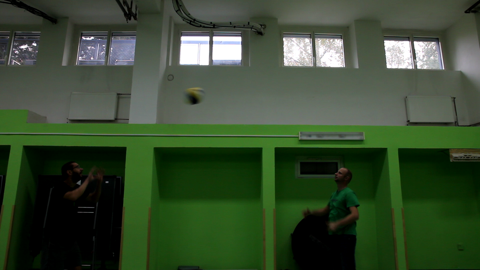}; 
\addplot [<-,>={Latex[length=0.8mm,width=0.5mm,angle'=25,open,round]},,domain=0:1,samples=10,style=semithick,color={rgb,255:red,48; green,255; blue,0}]({719.2974 + 12.7935*x + 2.7799*x^2},{400.1378 + -5.8888*x + 3.2376*x^2});  
\addplot [<-,>={Latex[length=1.1134mm,width=0.5mm,angle'=25,open,round]},,domain=0:1,samples=10,style=semithick,color={rgb,255:red,60; green,255; blue,0}]({668.3208 + 23.8892*x + 5.3404*x^2},{422.1211 + -26.1562*x + 11.7976*x^2});  
\addplot [<-,>={Latex[length=0.88288mm,width=0.5mm,angle'=25,open,round]},,domain=0:1,samples=10,style=semithick,color={rgb,255:red,88; green,255; blue,0}]({614.818 + 27.1363*x + -2.9061*x^2},{439.9486 + -5.0166*x + -5.1349*x^2});  
\addplot [<-,>={Latex[length=0.71654mm,width=0.5mm,angle'=25,open,round]},,domain=0:1,samples=10,style=semithick,color={rgb,255:red,128; green,255; blue,0}]({566.9877 + 21.8011*x + -0.39399*x^2},{465.4482 + 1.6606*x + -2.7726*x^2});  
\addplot [<-,>={Latex[length=0.80037mm,width=0.5mm,angle'=25,open,round]},,domain=0:1,samples=2,style=semithick,color={rgb,255:red,91; green,255; blue,0}]({517.4987 + 24.0103*x},{495.1633 + 0.18803*x});  
\addplot [<-,>={Latex[length=0.65mm,width=0.5mm,angle'=25,open,round]},,domain=0:1,samples=10,style=semithick,color={rgb,255:red,89; green,255; blue,0}]({560.7121 + -23.9591*x + -0.046212*x^2},{437.3264 + 15.5585*x + 1.6251*x^2});  
\addplot [->,>={Latex[length=1.0547mm,width=0.5mm,angle'=25,open,round]},,domain=0:1,samples=2,style=semithick,color={rgb,255:red,67; green,255; blue,0}]({580.9595 + 27.16*x},{425.2688 + -16.2321*x});  
\addplot [->,>={Latex[length=0.91877mm,width=0.5mm,angle'=25,open,round]},,domain=0:1,samples=2,style=semithick,color={rgb,255:red,107; green,255; blue,0}]({622.7975 + 25.1388*x},{399.6616 + -11.3032*x});  
\addplot [->,>={Latex[length=0.64946mm,width=0.5mm,angle'=25,open,round]},,domain=0:1,samples=2,style=semithick,color={rgb,255:red,90; green,255; blue,0}]({670.0642 + 18.2749*x},{378.3212 + -6.7563*x});  
\addplot [->,>={Latex[length=0.82446mm,width=0.5mm,angle'=25,open,round]},,domain=0:1,samples=2,style=semithick,color={rgb,255:red,110; green,255; blue,0}]({703.5396 + 23.8722*x},{366.1461 + -6.4712*x});  
\addplot [->,>={Latex[length=0.61476mm,width=0.5mm,angle'=25,open,round]},,domain=0:1,samples=2,style=semithick,color={rgb,255:red,107; green,255; blue,0}]({748.3282 + 18.2739*x},{352.7391 + -2.4898*x});  
\addplot [->,>={Latex[length=0.65061mm,width=0.5mm,angle'=25,open,round]},,domain=0:1,samples=2,style=semithick,color={rgb,255:red,91; green,255; blue,0}]({784.7207 + 19.3267*x},{348.0638 + -2.7277*x});  
\addplot [->,>={Latex[length=0.70098mm,width=0.5mm,angle'=25,open,round]},,domain=0:1,samples=2,style=semithick,color={rgb,255:red,85; green,255; blue,0}]({819.0056 + 21.0236*x},{346.2616 + 0.49641*x});  
\addplot [->,>={Latex[length=0.60817mm,width=0.5mm,angle'=25,open,round]},,domain=0:1,samples=2,style=semithick,color={rgb,255:red,95; green,255; blue,0}]({856.5789 + 17.9384*x},{349.0749 + 3.3314*x});  
\addplot [->,>={Latex[length=0.63217mm,width=0.5mm,angle'=25,open,round]},,domain=0:1,samples=2,style=semithick,color={rgb,255:red,76; green,255; blue,0}]({892.4762 + 18.1647*x},{357.0792 + 5.4511*x});  
\addplot [->,>={Latex[length=0.64994mm,width=0.5mm,angle'=25,open,round]},,domain=0:1,samples=2,style=semithick,color={rgb,255:red,116; green,255; blue,0}]({928.939 + 17.8995*x},{369.484 + 7.7326*x});  
\addplot [->,>={Latex[length=0.62647mm,width=0.5mm,angle'=25,open,round]},,domain=0:1,samples=2,style=semithick,color={rgb,255:red,129; green,255; blue,0}]({964.5839 + 16.5759*x},{386.8429 + 8.858*x});  
\addplot [->,>={Latex[length=0.65177mm,width=0.5mm,angle'=25,open,round]},,domain=0:1,samples=2,style=semithick,color={rgb,255:red,114; green,255; blue,0}]({1000.7703 + 16.2683*x},{409.0946 + 10.8474*x});  
\addplot [->,>={Latex[length=0.65077mm,width=0.5mm,angle'=25,open,round]},,domain=0:1,samples=2,style=semithick,color={rgb,255:red,96; green,255; blue,0}]({1037.1809 + 15.8188*x},{437.0586 + 11.4418*x});  
\addplot [->,>={Latex[length=0.6781mm,width=0.5mm,angle'=25,open,round]},,domain=0:1,samples=2,style=semithick,color={rgb,255:red,89; green,255; blue,0}]({1072.7713 + 15.493*x},{469.1812 + 13.1834*x});  
\addplot [->,>={Latex[length=0.57906mm,width=0.5mm,angle'=25,open,round]},,domain=0:1,samples=2,style=semithick,color={rgb,255:red,90; green,255; blue,0}]({1108.8725 + 13.7443*x},{508.0181 + 10.6244*x});  
\end{axis} 
\end{tikzpicture} 

  \noindent} & \resizebox {0.33\textwidth}{!} {\begin{tikzpicture} 
\begin{axis}[y dir=reverse, 
 xmin=1,xmax=1280, 
 ymin=1,ymax=720, 
 xticklabels = \empty, yticklabels = \empty, 
 grid=none, axis equal image] 
\addplot graphics[xmin=1,xmax=1280,ymin=1,ymax=720] {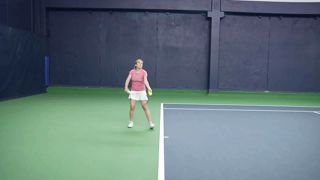}; 
\addplot [<-,>={Latex[length=0.75mm,width=0.5mm,angle'=25,open,round]},,domain=0:1,samples=10,style=semithick,color={rgb,255:red,141; green,255; blue,0}]({636.9313 + 0.1875*x + -1.2784*x^2},{156.8163 + 13.6738*x + -3.2894*x^2});  
\addplot [->,>={Latex[length=0.90056mm,width=0.5mm,angle'=25,open,round]},,domain=0:1,samples=2,style=semithick,color={rgb,255:red,202; green,255; blue,0}]({639.8156 + 0.22824*x},{136.9977 + -18.0098*x});  
\addplot [<-,>={Latex[length=0.54985mm,width=0.5mm,angle'=25,open,round]},,domain=0:1,samples=10,style=semithick,color={rgb,255:red,175; green,255; blue,0}]({643.0921 + -0.93863*x + 0.42565*x^2},{95.0099 + 10.9705*x + 0.011808*x^2});  
\addplot [->,>={Latex[length=0.50102mm,width=0.5mm,angle'=25,open,round]},,domain=0:1,samples=2,style=semithick,color={rgb,255:red,171; green,255; blue,0}]({644.9986 + 0.2308*x},{80 + -10.0177*x});  
\addplot [->,>={Latex[length=0.5mm,width=0.5mm,angle'=25,open,round]},,domain=0:1,samples=2,style=semithick,color={rgb,255:red,194; green,255; blue,0}]({648.0051 + -0.017475*x},{53 + -7*x});  
\addplot [->,>={Latex[length=0.5mm,width=0.5mm,angle'=25,open,round]},,domain=0:1,samples=2,style=semithick,color={rgb,255:red,197; green,255; blue,0}]({649.9205 + 1.2572*x},{32.9877 + -8.115*x});  
\addplot [->,>={Latex[length=0.5mm,width=0.5mm,angle'=25,open,round]},,domain=0:1,samples=2,style=semithick,color={rgb,255:red,187; green,255; blue,0}]({651.0424 + 4.9629*x},{9.5094 + 0.42809*x});  
\addplot [->,>={Latex[length=0.65mm,width=0.5mm,angle'=25,open,round]},,domain=0:1,samples=2,style=semithick,color={rgb,255:red,146; green,255; blue,0}]({689 + 0.5*x},{11 + 0.5*x});  
\addplot [<-,>={Latex[length=0.5mm,width=0.5mm,angle'=25,open,round]},,domain=0:1,samples=2,style=semithick,color={rgb,255:red,156; green,255; blue,0}]({691.0839 + 0.14932*x},{34.0015 + -7.9972*x});  
\addplot [<-,>={Latex[length=0.5mm,width=0.5mm,angle'=25,open,round]},,domain=0:1,samples=2,style=semithick,color={rgb,255:red,196; green,255; blue,0}]({692.1199 + -0.14379*x},{54.9975 + -6.997*x});  
\addplot [->,>={Latex[length=0.5mm,width=0.5mm,angle'=25,open,round]},,domain=0:1,samples=2,style=semithick,color={rgb,255:red,171; green,255; blue,0}]({693.5128 + -0.20619*x},{70.0133 + 7.9946*x});  
\addplot [->,>={Latex[length=1.5mm,width=0.5mm,angle'=25,open,round]},,domain=0:1,samples=10,style=semithick,color={rgb,255:red,176; green,255; blue,0}]({904.2614 + 49.2745*x + 31.4214*x^2},{78.7934 + 1.1248*x + -1.9966*x^2});  
\addplot [->,>={Latex[length=1.5mm,width=0.5mm,angle'=25,open,round]},,domain=0:1,samples=2,style=semithick,color={rgb,255:red,81; green,255; blue,0}]({1079.9997 + 89*x},{77.2226 + -0.015146*x});  
\addplot [->,>={Latex[length=0.5mm,width=0.5mm,angle'=25,open,round]},,domain=0:1,samples=2,style=semithick,color={rgb,255:red,151; green,255; blue,0}]({1272.9915 + 2.0339*x},{75.7691 + -0.074571*x});  
\end{axis} 
\end{tikzpicture} 

  \noindent} \\
\resizebox {0.33\textwidth}{!} {\begin{tikzpicture} 
\begin{axis}[y dir=reverse, 
 xmin=1,xmax=1920, 
 ymin=1,ymax=1080, 
 xticklabels = \empty, yticklabels = \empty, 
 grid=none, axis equal image] 
\addplot graphics[xmin=1,xmax=1920,ymin=1,ymax=1080] {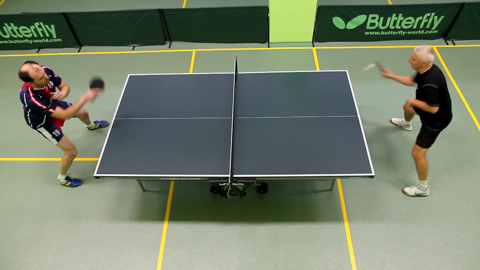}; 
\addplot [->,>={Latex[length=0.80233mm,width=0.5mm,angle'=25,open,round]},,domain=0:1,samples=2,style=semithick,color={rgb,255:red,255; green,192; blue,0}]({1457.9813 + 14.6466*x},{279.9582 + -6.5552*x});  
\addplot [->,>={Latex[length=0.82765mm,width=0.5mm,angle'=25,open,round]},,domain=0:1,samples=2,style=semithick,color=red]({1492.3889 + 15.0016*x},{277.1662 + 6.9966*x});  
\addplot [<-,>={Latex[length=1.5mm,width=0.5mm,angle'=25,open,round]},,domain=0:1,samples=2,style=semithick,color={rgb,255:red,36; green,255; blue,0}]({1227.752 + 72.693*x},{309.5445 + -16.93*x});  
\addplot [>={Latex[length=1.5mm,width=0.5mm,angle'=25,open,round]},,domain=0:1,samples=2,style=semithick,color={rgb,255:red,36; green,255; blue,0}]({1300.4449 + 17.3109*x},{292.6145 + -0.86501*x});  
\addplot [<-,>={Latex[length=1.5mm,width=0.5mm,angle'=25,open,round]},,domain=0:1,samples=2,style=semithick,color={rgb,255:red,137; green,255; blue,0}]({1096.5082 + 97.3548*x},{346.2751 + -27.7555*x});  
\addplot [<-,>={Latex[length=1.5mm,width=0.5mm,angle'=25,open,round]},,domain=0:1,samples=2,style=semithick,color={rgb,255:red,171; green,255; blue,0}]({970.7656 + 95.2787*x},{389.3272 + -33.2001*x});  
\addplot [<-,>={Latex[length=1.5mm,width=0.5mm,angle'=25,open,round]},,domain=0:1,samples=2,style=semithick,color={rgb,255:red,154; green,255; blue,0}]({851.1877 + 73.7748*x},{437.4533 + -30.544*x});  
\addplot [<-,>={Latex[length=1.5mm,width=0.5mm,angle'=25,open,round]},,domain=0:1,samples=2,style=semithick,color={rgb,255:red,174; green,255; blue,0}]({734.7472 + 86.0782*x},{491.4671 + -40.8352*x});  
\addplot [<-,>={Latex[length=1.5mm,width=0.5mm,angle'=25,open,round]},,domain=0:1,samples=2,style=semithick,color={rgb,255:red,202; green,255; blue,0}]({627.8199 + 80.156*x},{548.6697 + -43.714*x});  
\addplot [<-,>={Latex[length=1.5mm,width=0.5mm,angle'=25,open,round]},,domain=0:1,samples=2,style=semithick,color={rgb,255:red,191; green,255; blue,0}]({540.0152 + 64.7365*x},{601.6812 + -39.0923*x});  
\addplot [<-,>={Latex[length=1.5mm,width=0.5mm,angle'=25,open,round]},,domain=0:1,samples=2,style=semithick,color={rgb,255:red,197; green,255; blue,0}]({454.9681 + 50.021*x},{605.5704 + -3.717*x});  
\addplot [<-,>={Latex[length=1.5mm,width=0.5mm,angle'=25,open,round]},,domain=0:1,samples=2,style=semithick,color={rgb,255:red,221; green,255; blue,0}]({354.8783 + 39.1734*x},{613.9369 + -4.4856*x});  
\addplot [,domain=0:1,samples=2,style=semithick,color={rgb,255:red,155; green,255; blue,0}]({424.5583 + -55.7686*x},{555.5719 + 20.5543*x});  
\addplot [,domain=0:1,samples=2,style=semithick,color={rgb,255:red,155; green,255; blue,0}]({368.7897 + -51.9508*x},{576.1262 + 20.3527*x});  
\addplot [>={Latex[length=1.5mm,width=0.5mm,angle'=25,open,round]},,domain=0:1,samples=2,style=semithick,color={rgb,255:red,66; green,255; blue,0}]({506.8662 + 75.187*x},{525.9453 + -24.9894*x});  
\addplot [->,>={Latex[length=1.5mm,width=0.5mm,angle'=25,open,round]},,domain=0:1,samples=2,style=semithick,color={rgb,255:red,66; green,255; blue,0}]({582.0532 + 48.7809*x},{500.9558 + -15.3057*x});  
\addplot [->,>={Latex[length=1.5mm,width=0.5mm,angle'=25,open,round]},,domain=0:1,samples=2,style=semithick,color={rgb,255:red,124; green,255; blue,0}]({679.0597 + 124.617*x},{470.2103 + -35.3502*x});  
\addplot [>={Latex[length=1.5mm,width=0.5mm,angle'=25,open,round]},,domain=0:1,samples=2,style=semithick,color={rgb,255:red,54; green,255; blue,0}]({996.0197 + 48.5325*x},{389.6903 + -9.4011*x});  
\addplot [->,>={Latex[length=1.5mm,width=0.5mm,angle'=25,open,round]},,domain=0:1,samples=2,style=semithick,color={rgb,255:red,54; green,255; blue,0}]({1044.5521 + 44.1721*x},{380.2891 + -7.6613*x});  
\addplot [->,>={Latex[length=1.5mm,width=0.5mm,angle'=25,open,round]},,domain=0:1,samples=2,style=semithick,color={rgb,255:red,198; green,255; blue,0}]({1126.9204 + 87.9332*x},{366.3918 + -11.5106*x});  
\addplot [->,>={Latex[length=1.5mm,width=0.5mm,angle'=25,open,round]},,domain=0:1,samples=2,style=semithick,color={rgb,255:red,200; green,255; blue,0}]({1243.9301 + 78.0908*x},{352.0538 + -5.7717*x});  
\addplot [->,>={Latex[length=1.5mm,width=0.5mm,angle'=25,open,round]},,domain=0:1,samples=10,style=semithick,color={rgb,255:red,191; green,255; blue,0}]({1347.102 + 34.2644*x + 19.6504*x^2},{343.7901 + 16.614*x + -36.3724*x^2});  
\addplot [->,>={Latex[length=1.5mm,width=0.5mm,angle'=25,open,round]},,domain=0:1,samples=2,style=semithick,color={rgb,255:red,221; green,255; blue,0}]({1450.2535 + 64.3845*x},{299.5219 + -31.2673*x});  
\end{axis} 
\end{tikzpicture} 

  \noindent} & \resizebox {0.33\textwidth}{!} {\begin{tikzpicture} 
\begin{axis}[y dir=reverse, 
 xmin=1,xmax=1920, 
 ymin=1,ymax=1080, 
 xticklabels = \empty, yticklabels = \empty, 
 grid=none, axis equal image] 
\addplot graphics[xmin=1,xmax=1920,ymin=1,ymax=1080] {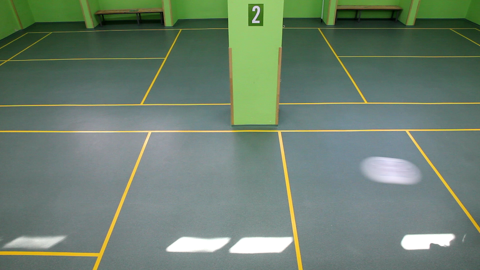}; 
\addplot [->,>={Latex[length=1.5mm,width=0.5mm,angle'=25,open,round]},,domain=0:1,samples=10,style=semithick,color={rgb,255:red,36; green,255; blue,0}]({1620.9035 + -89.6051*x + -20.998*x^2},{691.4233 + -10.3668*x + -1.5298*x^2});  
\addplot [<-,>={Latex[length=1.4977mm,width=0.5mm,angle'=25,open,round]},,domain=0:1,samples=2,style=semithick,color={rgb,255:red,93; green,255; blue,0}]({1255.3764 + 149.5484*x},{668.1502 + 8.218*x});  
\addplot [<-,>={Latex[length=1.5mm,width=0.5mm,angle'=25,open,round]},,domain=0:1,samples=2,style=semithick,color={rgb,255:red,103; green,255; blue,0}]({1009.8496 + 160.2198*x},{658.2604 + 2.9243*x});  
\addplot [<-,>={Latex[length=1.5mm,width=0.5mm,angle'=25,open,round]},,domain=0:1,samples=2,style=semithick,color={rgb,255:red,120; green,255; blue,0}]({754.8126 + 160.1506*x},{650.0518 + 5.9402*x});  
\addplot [<-,>={Latex[length=1.5mm,width=0.5mm,angle'=25,open,round]},,domain=0:1,samples=10,style=semithick,color={rgb,255:red,123; green,255; blue,0}]({504.9771 + 161.2545*x + -1.1756*x^2},{648.6529 + 1.2582*x + -2.818*x^2});  
\addplot [<-,>={Latex[length=1.5mm,width=0.5mm,angle'=25,open,round]},,domain=0:1,samples=2,style=semithick,color={rgb,255:red,139; green,255; blue,0}]({285.0022 + 154.9979*x},{645.324 + -0.063209*x});  
\addplot [<-,>={Latex[length=1.498mm,width=0.5mm,angle'=25,open,round]},,domain=0:1,samples=10,style=semithick,color={rgb,255:red,144; green,255; blue,0}]({70.319 + 146.5621*x + 3.1168*x^2},{649.0555 + -11.8485*x + 7.6674*x^2});  
\addplot [->,>={Latex[length=1.5mm,width=0.5mm,angle'=25,open,round]},,domain=0:1,samples=10,style=semithick,color={rgb,255:red,34; green,255; blue,0}]({1630.0175 + -134.4481*x + 5.9237*x^2},{658.5887 + -23.8713*x + 0.69414*x^2});  
\addplot [<-,>={Latex[length=1.5mm,width=0.5mm,angle'=25,open,round]},,domain=0:1,samples=2,style=semithick,color={rgb,255:red,48; green,255; blue,0}]({1245.047 + 164.8531*x},{599.6336 + 21.1452*x});  
\addplot [<-,>={Latex[length=1.5mm,width=0.5mm,angle'=25,open,round]},,domain=0:1,samples=10,style=semithick,color={rgb,255:red,144; green,255; blue,0}]({973.9368 + 173.6804*x + 7.3673*x^2},{553.3733 + 56.9709*x + -28.9639*x^2});  
\addplot [<-,>={Latex[length=1.5mm,width=0.5mm,angle'=25,open,round]},,domain=0:1,samples=2,style=semithick,color={rgb,255:red,94; green,255; blue,0}]({745.2943 + 169.6781*x},{543.2435 + 7.39*x});  
\addplot [<-,>={Latex[length=1.5mm,width=0.5mm,angle'=25,open,round]},,domain=0:1,samples=2,style=semithick,color={rgb,255:red,151; green,255; blue,0}]({505.3044 + 179.6558*x},{528.0451 + 7.8634*x});  
\addplot [<-,>={Latex[length=1.5mm,width=0.5mm,angle'=25,open,round]},,domain=0:1,samples=2,style=semithick,color={rgb,255:red,118; green,255; blue,0}]({300.0003 + 159.9999*x},{515.2918 + -0.14772*x});  
\addplot [<-,>={Latex[length=1.5mm,width=0.5mm,angle'=25,open,round]},,domain=0:1,samples=2,style=semithick,color={rgb,255:red,177; green,255; blue,0}]({80.4165 + 164.5175*x},{507.3096 + 8.9093*x});  
\addplot [->,>={Latex[length=0.5mm,width=0.5mm,angle'=25,open,round]},,domain=0:1,samples=2,style=semithick,color={rgb,255:red,202; green,255; blue,0}]({49.7732 + -0.1674*x},{500.0025 + -14.9981*x});  
\end{axis} 
\end{tikzpicture} 

  \noindent} & \resizebox {0.33\textwidth}{!} {\begin{tikzpicture} 
\begin{axis}[y dir=reverse, 
 xmin=1,xmax=1920, 
 ymin=1,ymax=1080, 
 xticklabels = \empty, yticklabels = \empty, 
 grid=none, axis equal image] 
\addplot graphics[xmin=1,xmax=1920,ymin=1,ymax=1080] {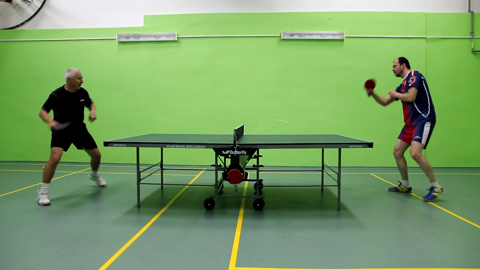}; 
\addplot [<-,>={Latex[length=1.5mm,width=0.5mm,angle'=25,open,round]},,domain=0:1,samples=2,style=semithick,color={rgb,255:red,87; green,255; blue,0}]({1107.146 + 20.1875*x},{479.8229 + 3.9753*x});  
\addplot [>={Latex[length=1.5mm,width=0.5mm,angle'=25,open,round]},,domain=0:1,samples=2,style=semithick,color={rgb,255:red,87; green,255; blue,0}]({1127.3335 + 22.5604*x},{483.7981 + 4.3971*x});  
\addplot [<-,>={Latex[length=1.5mm,width=0.5mm,angle'=25,open,round]},,domain=0:1,samples=2,style=semithick,color={rgb,255:red,166; green,255; blue,0}]({1034.0485 + 47.9774*x},{470.553 + 5.2083*x});  
\addplot [<-,>={Latex[length=1.5mm,width=0.5mm,angle'=25,open,round]},,domain=0:1,samples=2,style=semithick,color={rgb,255:red,165; green,255; blue,0}]({962.0045 + 47.9911*x},{467.8362 + 1.3229*x});  
\addplot [<-,>={Latex[length=1.5mm,width=0.5mm,angle'=25,open,round]},,domain=0:1,samples=2,style=semithick,color={rgb,255:red,219; green,255; blue,0}]({891.9961 + 47.0191*x},{470.9317 + -2.6624*x});  
\addplot [<-,>={Latex[length=1.5mm,width=0.5mm,angle'=25,open,round]},,domain=0:1,samples=10,style=semithick,color={rgb,255:red,227; green,255; blue,0}]({813.3508 + 27.7166*x + 32.9718*x^2},{486.6271 + -15.7955*x + 6.5432*x^2});  
\addplot [<-,>={Latex[length=1.5mm,width=0.5mm,angle'=25,open,round]},,domain=0:1,samples=2,style=semithick,color={rgb,255:red,171; green,255; blue,0}]({752.7682 + 21.5969*x},{496.1826 + -5.8089*x});  
\addplot [>={Latex[length=1.5mm,width=0.5mm,angle'=25,open,round]},,domain=0:1,samples=2,style=semithick,color={rgb,255:red,171; green,255; blue,0}]({774.3651 + 21.5898*x},{490.3737 + -4.6211*x});  
\addplot [<-,>={Latex[length=1.5mm,width=0.5mm,angle'=25,open,round]},,domain=0:1,samples=2,style=semithick,color={rgb,255:red,168; green,255; blue,0}]({681.8234 + 46.1615*x},{519.4747 + -15.5196*x});  
\addplot [<-,>={Latex[length=1.5mm,width=0.5mm,angle'=25,open,round]},,domain=0:1,samples=10,style=semithick,color={rgb,255:red,188; green,255; blue,0}]({614.2393 + 22.125*x + 24.8971*x^2},{536.8671 + 25.5266*x + -35.5609*x^2});  
\addplot [,domain=0:1,samples=10,style=semithick,color={rgb,255:red,126; green,255; blue,0}]({540.1392 + 19.2405*x + 29.2398*x^2},{512.2033 + 4.8735*x + 10.1709*x^2});  
\addplot [,domain=0:1,samples=10,style=semithick,color={rgb,255:red,166; green,255; blue,0}]({517.3192 + -18.0499*x + -29.1555*x^2},{505.8799 + -6.4384*x + -5.0294*x^2});  
\addplot [,domain=0:1,samples=10,style=semithick,color={rgb,255:red,82; green,255; blue,0}]({499.8924 + 98.0447*x + -4.6225*x^2},{462.3037 + -8.3559*x + 1.0919*x^2});  
\addplot [,domain=0:1,samples=2,style=semithick,color={rgb,255:red,105; green,255; blue,0}]({637.0598 + 18.3908*x},{453.0351 + -0.040156*x});  
\addplot [,domain=0:1,samples=2,style=semithick,color={rgb,255:red,105; green,255; blue,0}]({655.4506 + 67.2668*x},{452.9949 + -0.067039*x});  
\addplot [,domain=0:1,samples=2,style=semithick,color={rgb,255:red,80; green,255; blue,0}]({841.7496 + -36.23*x},{458.3392 + -2.9852*x});  
\addplot [,domain=0:1,samples=2,style=semithick,color={rgb,255:red,80; green,255; blue,0}]({805.5196 + -34.4941*x},{455.354 + -0.56207*x});  
\addplot [,domain=0:1,samples=10,style=semithick,color={rgb,255:red,110; green,255; blue,0}]({883.0739 + 64.7957*x + 3.896*x^2},{462.829 + 5.1487*x + 3.7284*x^2});  
\addplot [>={Latex[length=1.5mm,width=0.5mm,angle'=25,open,round]},,domain=0:1,samples=2,style=semithick,color={rgb,255:red,82; green,255; blue,0}]({986.3582 + 31.0845*x},{477.5472 + 5.8027*x});  
\addplot [->,>={Latex[length=1.5mm,width=0.5mm,angle'=25,open,round]},,domain=0:1,samples=2,style=semithick,color={rgb,255:red,82; green,255; blue,0}]({1017.4426 + 32.5581*x},{483.3499 + 6.89*x});  
\addplot [->,>={Latex[length=1.5mm,width=0.5mm,angle'=25,open,round]},,domain=0:1,samples=2,style=semithick,color={rgb,255:red,170; green,255; blue,0}]({1083.9701 + 56.0858*x},{499.1068 + 15.6932*x});  
\addplot [>={Latex[length=1.5mm,width=0.5mm,angle'=25,open,round]},,domain=0:1,samples=2,style=semithick,color={rgb,255:red,134; green,255; blue,0}]({1171.8649 + 30.8048*x},{526.0386 + 7.8655*x});  
\addplot [->,>={Latex[length=1.5mm,width=0.5mm,angle'=25,open,round]},,domain=0:1,samples=2,style=semithick,color={rgb,255:red,134; green,255; blue,0}]({1202.6697 + 20.9115*x},{533.9041 + -8.2387*x});  
\addplot [->,>={Latex[length=1.5mm,width=0.5mm,angle'=25,open,round]},,domain=0:1,samples=2,style=semithick,color={rgb,255:red,192; green,255; blue,0}]({1248.8255 + 49.0435*x},{516.4605 + -15.8655*x});  
\addplot [->,>={Latex[length=1.5mm,width=0.5mm,angle'=25,open,round]},,domain=0:1,samples=2,style=semithick,color={rgb,255:red,160; green,255; blue,0}]({1320.7678 + 44.3804*x},{494.0063 + -10.3723*x});  
\addplot [->,>={Latex[length=1.5mm,width=0.5mm,angle'=25,open,round]},,domain=0:1,samples=2,style=semithick,color={rgb,255:red,158; green,255; blue,0}]({1387.8887 + 43.0712*x},{479.2661 + -6.5305*x});  
\addplot [->,>={Latex[length=1.5mm,width=0.5mm,angle'=25,open,round]},,domain=0:1,samples=2,style=semithick,color={rgb,255:red,147; green,255; blue,0}]({1449.0353 + 40.9435*x},{471.5125 + -2.8207*x});  
\end{axis} 
\end{tikzpicture} 

  \noindent} \\
\end{tabular}
\caption{Trajectory recovery for selected sequences from the FMO dataset~\cite{fmo}. Intersection over Union (IoU) with the ground truth occupancy mask is color coded using the scale from Figure~\ref{tbl:tbd_imgs}. Arrows indicate the direction of the motion.}
\label{tbl:fmo_imgs}
\end{figure*}
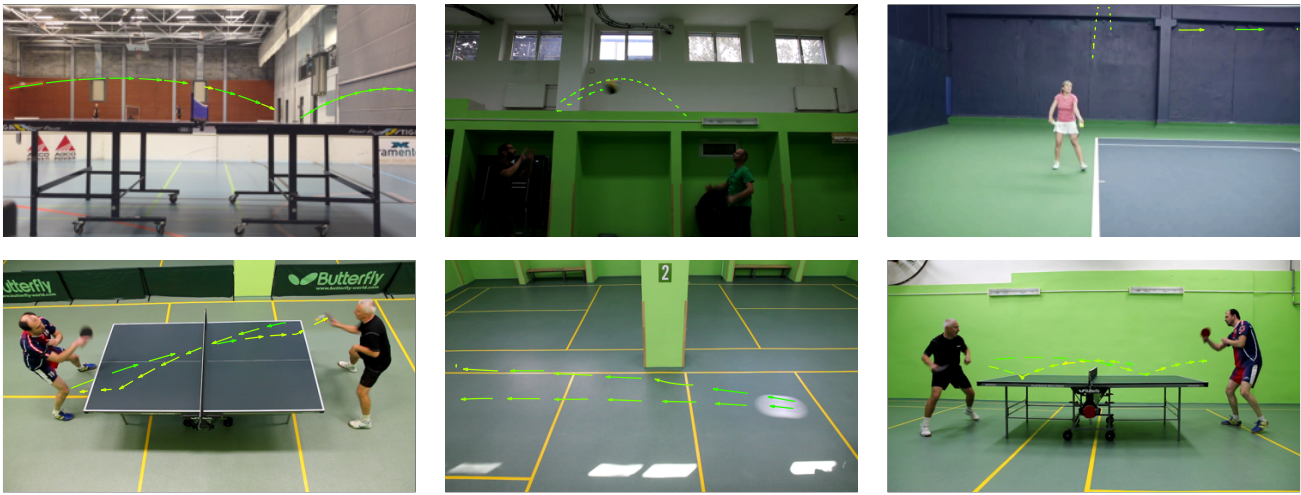

\begin{figure}
	\noindent\begin{minipage}[t]{.24\linewidth}
		\centering
		\includegraphics[width=\textwidth]{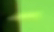}
	\end{minipage}\hfill%
	\begin{minipage}[t]{.24\linewidth}
		\centering
		\includegraphics[width=\textwidth]{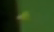}
	\end{minipage}\hfill%
	\begin{minipage}[t]{.24\linewidth}
		\centering
		\includegraphics[width=\textwidth]{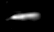}
	\end{minipage}\hfill%
	\begin{minipage}[t]{.12\linewidth}
		\centering
		\includegraphics[width=\textwidth]{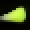}
	\end{minipage}\hfill%
	\begin{minipage}[t]{.12\linewidth}
		\centering
		\includegraphics[width=\textwidth]{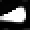}
	\end{minipage}
	\begin{minipage}[t]{.24\linewidth}
		\centering
		\includegraphics[width=\textwidth]{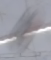}
	\end{minipage}\hfill%
	\begin{minipage}[t]{.24\linewidth}
		\centering
		\includegraphics[width=\textwidth]{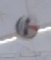}
	\end{minipage}\hfill%
	\begin{minipage}[t]{.24\linewidth}
		\centering
		\includegraphics[width=\textwidth]{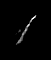}
	\end{minipage}\hfill%
	\begin{minipage}[t]{.12\linewidth}
		\centering
		\includegraphics[width=\textwidth]{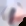}
	\end{minipage}\hfill%
	\begin{minipage}[t]{.12\linewidth}
		\centering
		\includegraphics[width=\textwidth]{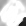}
	\end{minipage}
	\begin{minipage}[t]{.24\linewidth}
		\centering
		Input $I$
	\end{minipage}\hfill%
	\begin{minipage}[t]{.24\linewidth}
		\centering
		High FPS%
	\end{minipage}\hfill%
	\begin{minipage}[t]{.24\linewidth}
		\centering
		Blur $H$
	\end{minipage}\hfill%
	\begin{minipage}[t]{.12\linewidth}
		\centering
		$F$
	\end{minipage}\hfill%
	\begin{minipage}[t]{.12\linewidth}
		\centering
		$M$
	\end{minipage}%
	\caption{Deblatting examples. From left to right: the input image, corresponding high-speed camera frame, estimated blur $H$, estimated appearance $F$ and shape $M$.}
	\label{fig:deblurring_FMH}
\end{figure}
\begin{figure}
	\noindent\begin{minipage}[t]{.24\linewidth}
		\centering
		\includegraphics[width=\textwidth]{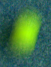}
	\end{minipage}\hfill%
	\begin{minipage}[t]{.24\linewidth}
		\centering
		\includegraphics[width=\textwidth]{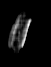}
	\end{minipage}\hfill%
	\begin{minipage}[t]{.24\linewidth}
		\centering
		\setlength{\fboxsep}{0pt}
		\includegraphics[width=\textwidth]{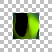}
	\end{minipage}\hfill%
	\begin{minipage}[t]{.24\linewidth}
		\centering
		\setlength{\fboxsep}{0pt}
		\includegraphics[width=\textwidth]{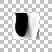}
	\end{minipage}
	\begin{minipage}[t]{.24\linewidth}
		\centering
		\includegraphics[width=\textwidth]{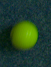}
	\end{minipage}\hfill%
	\begin{minipage}[t]{.24\linewidth}
		\centering
		\includegraphics[width=\textwidth]{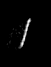}
	\end{minipage}\hfill%
	\begin{minipage}[t]{.24\linewidth}
		\centering
		\includegraphics[width=\textwidth]{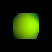}
	\end{minipage}\hfill%
	\begin{minipage}[t]{.24\linewidth}
		\centering
		\includegraphics[width=\textwidth]{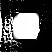}
	\end{minipage}
	\begin{minipage}[t]{.24\linewidth}
		\centering\scriptsize
		\mbox{$I$ (top)/high FPS}
	\end{minipage}\hfill%
	\begin{minipage}[t]{.24\linewidth}
		\centering
		$H$
	\end{minipage}\hfill%
	\begin{minipage}[t]{.24\linewidth}
		\centering
		$F$
	\end{minipage}\hfill%
	\begin{minipage}[t]{.24\linewidth}
		\centering
		$M$
	\end{minipage}%
	\caption{Shadow and blur estimation. Top: the domain of $F$ is set too small and the shadow causes artifacts in $H$. Bottom: the domain of $F$ is larger, $M$ can compensate for the shadow and the blur $H$ is estimated correctly.}
	\label{fig:deblurring_shadow}
\end{figure}

\section{Experiments}
\label{sec:experiments}
We show the results of Tracking by Deblatting and compare it with other trackers on the task of long-term tracking of motion-blurred objects in real-life video sequences. As a baseline, we chose the FMO detector (FMOd, \cite{fmo}), specifically proposed for tracking fast moving objects, and the Discriminative Correlation Filter with Channel and Spatial Reliability (CSR-DCF, \cite{csrdcf}), which performs well on standard benchmarks such as VOT~\cite{vot2016}. CSR-DCF was not designed to track objects undergoing large changes in velocity within a single sequence and would perform poorly in the comparison. We therefore augmented CSR-DCF by FMOd reinitialization every time it outputs the same bounding box in consecutive frames, which is considered a fail. We use FMOd for automatic initialization of both TbD and CSR-DCF to avoid manual input and we skip the first two frames of every sequence to establish background $B$ and initialize CSR-DCF. The rest of the sequence is processed causally, $B$ is estimated as a moving median of the  past 3~--~5 frames.

The goal of TbD is to produce a precise intra-frame motion trajectory, not only a single position per frame in the form of a bounding box. Fig. \ref{fig:fitting} shows examples of trajectory estimation.
The left column is the input image with the estimated PSF superimposed in white and the right column shows the estimated motion trajectory. The efficacy of trajectory fitting is a crucial part of the framework, the estimated blur can contain various artifacts (\eg~in the top example due to the ball shadow) and the trajectory fit still recovers the actual motion.

The comparison with baseline methods was conducted on a new dataset consisting of 12 sequences with different objects in motion and setting (different kinds of sports, objects in flight or rolled on the ground, indoor/outdoor). The sequences contain abrupt changes of motion, such as bounces and interactions with players, and a wide range of speeds.
The dataset is annotated with the ground-truth trajectory for each frame, obtained from  a high-speed camera footage.
We compare the method performance in predicting the motion trajectory in each frame. We therefore generalize IoU, the standard measure of position accuracy, to trajectories and define a new measure \emph{Trajectory-IoU} (TIoU):
\begin{equation}
	\label{eq:exp_TIoU_def}
	\operatorname{TIoU}(\mathcal{C},\mathcal{C}^*; M^*) = \int_{t} \operatorname{IoU}\left(\rule{0pt}{2ex}M^*_{\mathcal{C}(t)},\, M^*_{\mathcal{C}^*(t)}\right)\mathrm{d}t,
\end{equation}
where $\mathcal{C}$ is the predicted trajectory, $\mathcal{C}^*$ is the ground-truth trajectory, $M^*$ is a disk mask with true object diameter obtained from the ground truth, and $M_x$ denotes $M$ placed at location $x$. TIoU can be regarded as the standard IoU averaged over each position on the estimated trajectory. In practice, we discretize the exposure time into evenly spaced timestamps and calculate IoU of the ground-truth object location and prediction by the tracker at the timestamps and average these measurements. CSR-DCF tracker only outputs positions, so in this case we estimate linear trajectories from positions in neighboring frames and then calculate TIoU.

\begin{table*}
\begin{center}
\begin{tabular}{l|r|c|c|c|c|c|c|c|c|c|c|c}
\hline
\multirow{2}{*}{Sequence name} & \multirow{2}{*}{\#} & \multicolumn{2}{c|}{CSR-DCF~\cite{csrdcf}}  & \multicolumn{2}{c|}{FMO~\cite{fmo}} & \multicolumn{2}{c|}{TbD-T0, 0} & \multicolumn{2}{c|}{TbD-T0, 0.5} & \multicolumn{2}{c|}{TbD-T1, 1} & TbD-O \\  \cline{3-13}
 & & TIoU & Rcl & TIoU & Rcl & TIoU & Rcl & TIoU & Rcl & TIoU & Rcl & TIoU \\ \hline

badminton\_white & 40 & .275 & 0.39 & .242 & 0.34 & .673 & 0.92 & .674 & \textcolor{cyan}{0.95} & \textcolor{blue}{.711} & \textcolor{cyan}{0.95} & .792\\  
 badminton\_yellow & 57 & .047 & 0.11 & .236 & 0.31 & .615 & \textcolor{cyan}{0.89} & .623 & \textcolor{cyan}{0.89} & \textcolor{blue}{.633} & 0.85 & .788\\  
 pingpong & 58 & .060 & 0.14 & .064 & 0.12 & .583 & 0.89 & \textcolor{blue}{.587} & 0.89 & .536 & \textcolor{cyan}{0.91} & .697\\  
 tennis & 38 & .249 & 0.83 & .596 & 0.78 & .577 & 0.81 & .573 & 0.81 & \textcolor{blue}{.633} & \textcolor{cyan}{0.86} & .827\\  
 volleyball & 41 & .373 & 0.69 & .537 & 0.72 & .552 & 0.87 & .587 & 0.90 & \textcolor{blue}{.741} & \textcolor{cyan}{0.92} & .836\\  
 throw\_floor & 40 & .262 & 0.74 & .272 & 0.37 & .746 & \textcolor{cyan}{1.00} & .768 & \textcolor{cyan}{1.00} & \textcolor{blue}{.817} & \textcolor{cyan}{1.00} & .864\\  
 throw\_soft & 60 & .470 & 0.93 & .377 & 0.57 & .585 & 0.90 & .539 & 0.90 & \textcolor{blue}{.641} & \textcolor{cyan}{0.95} & .707\\  
 throw\_tennis & 45 & .347 & 0.91 & .507 & 0.65 & .688 & \textcolor{cyan}{1.00} & .781 & \textcolor{cyan}{1.00} & \textcolor{blue}{.852} & \textcolor{cyan}{1.00} & .872\\  
 roll\_golf & 16 & .406 & \textcolor{cyan}{1.00} & .187 & 0.71 & .414 & \textcolor{cyan}{1.00} & .346 & \textcolor{cyan}{1.00} & \textcolor{blue}{.851} & \textcolor{cyan}{1.00} & .898\\  
 fall\_cube & 20 & .422 & \textcolor{cyan}{0.89} & .408 & 0.78 & .553 & \textcolor{cyan}{0.89} & .669 & \textcolor{cyan}{0.89} & \textcolor{blue}{.704} & \textcolor{cyan}{0.89} & .744\\  
 hit\_tennis & 30 & .316 & \textcolor{cyan}{0.93} & .381 & 0.68 & .564 & \textcolor{cyan}{0.93} & .570 & \textcolor{cyan}{0.93} & \textcolor{blue}{.662} & \textcolor{cyan}{0.93} & .828\\  
 hit\_tennis2 & 26 & .289 & 0.79 & .414 & 0.71 & .459 & \textcolor{cyan}{0.83} & .493 & \textcolor{cyan}{0.83} & \textcolor{blue}{.627} & \textcolor{cyan}{0.83} & .738\\  
 \hline 
  Average  & 39 & .293 & 0.70 & .352 & 0.56 & .584 & 0.91 & .601 & 0.92 & \textcolor{blue}{.701} & \textcolor{cyan}{0.93} & .799\\

\hline
\end{tabular}
\end{center}
\caption{Trajectory Intersection over Union (TIoU) and Recall (Rcl) on the TbD dataset -- 
comparison  of the TbD, CSR-DCF\cite{csrdcf} trackers and the Fast Moving Object method~\cite{fmo}. 
CSR-DCF is a standard,  well-performning~\cite{vot2018}, near-real time tracker.
TbD tracker settings: TbD without template and with exponential forgetting factors~\eqref{eq:update_model} $\gamma=0$ (TbD-T0, 0) and $\gamma=0.5$ (TbD-T0, 0.5), TbD with template and $\gamma=1$ (TbD-T1, 1), TbD with oracle (TbD-O). The highest TIoU for each sequence is highlighted in blue color and the highest recall in cyan color. TbD-O shows the highest attainable TIoU for TbD as a reference point when predictions are precise. The number of frames is indicated by \#.}
\label{tbl:tbd}
\end{table*}

The results of the comparison are presented in Table~\ref{tbl:tbd}. We evaluated three flavors of TbD that differ in the presence of the initial user-supplied template $\hat F$ and the learning rate $\gamma$ of the object model in \eqref{eq:update_model}. The presented flavors are:
\vspace*{-1ex}
\begin{itemize}[leftmargin=*]
	\setlength\itemsep{-0.25em}
	\item TbD-T0,0: Object template not available, model update is instantaneous (memory-less), $\gamma=0$.
	\item TbD-T0,0.5: Object template not available, model is updated with the learning rate $\gamma=0.5$.
	\item TbD-T1,1: Object template available, model remains constant and equal to the template, $\gamma=1$.
\end{itemize}
\vspace*{-1ex}
The TbD outperforms baseline methods on average by a wide margin, both in the traditional recall measure (a detection is called true positive if it overlaps with the ground truth) as well as in trajectory accuracy TIoU. FMOd is less accurate and more prone to false positives as it lacks any prediction step and by design ignores slow objects. CSR-DCF, despite reinitializations by FMOd, fails to detect fast moving objects accurately. Among TbD flavors, it is no surprise that availability of the object template is beneficial and outperforms other versions. However, even if the template is not available, TbD can learn the object model and updating the appearance model gradually during tracking is preferable to instantaneous updates.

To evaluate the performance of the core part of TbD that consists of deblatting and trajectory fitting alone, we provide results of a special version of the proposed method called ``TbD with oracle'', TbD-O. This behaves like regular TbD but with a perfect trajectory prediction step. We use the ground-truth trajectory to supply the region $D$ to the deblatting step exactly as if it were predicted by the prediction step, effectively bypassing the long-term tracking logic of TbD. The rest is identical to TbD-T1,1. TbD with oracle tests the performance and potential of the deblatting and trajectory estimation alone because failures do not cause long-term damage~--~success in one frame is independent of success in the previous frame.

\begin{table}
\begin{center}
\begin{tabular}{l|r|r|r|r}
\hline
\multirow{2}{*}{FMO dataset}  &  \multicolumn{2}{c|}{FMO~\cite{fmo}} &  \multicolumn{2}{c}{TbD-T0, 0.5}  \\  \cline{2-5}
 &  Prec.  & Recall  & Prec. & Recall  \\ 
\hline
Average &  59.2 & 35.5 & \textbf{81.6} & \textbf{41.1} \\
\hline
\end{tabular}
\end{center}
\caption{Precision and recall of the TbD tracker (setting: TbD without template and with exponential forgetting factor~\eqref{eq:update_model} $\gamma=0.5$) and the FMO method~\cite{fmo}, average on the 16 sequences of the FMO~dataset. 
}
\label{tbl:fmo}
\end{table}

Table~\ref{tbl:fmo} shows aggregated results for the FMO dataset \cite{fmo}. This dataset does not contain ground-truth trajectory data, we therefore report traditional precision/recall measure, which is derived from the detection and ground-truth bounding-box IoU. On this dataset, the proposed TbD method is slightly better in recall, owing to the fact that initial detection is done by FMOd and if FMOd fails then TbD cannot start the tracking, but significantly better in terms of precision. Results on individual sequences are in the supplementary.

A visual demonstration of the tracking by the proposed method on the TbD dataset and the FMO dataset is shown in Figs.~\ref{tbl:tbd_imgs} and \ref{tbl:fmo_imgs}.
Each image shows results of tracking in one sequence from the evaluation dataset superimposed on a single image from the sequence. Arrows depict trajectories detected in a particular frame and the color encodes the corresponding TIoU from green=1 to red=0 (false positive). We can see that the trajectory is estimated successfully with the exception of frames where the object is in direct contact with other moving objects, which throws off the local estimation of background.

Examples of the deblatting alone are in Figs.~\ref{fig:deblurring_FMH} and \ref{fig:deblurring_shadow}. Fig.~\ref{fig:deblurring_FMH} contains from left to right the input frame (crop), corresponding frame from the high-speed camera, estimated motion PSF $H$, estimated object $F$ and object shape $M$. In the top row, we see that the shape of the badminton shuttlecock, though not circular, is estimated correctly. In the bottom row, we see that if the non-uniform object undergoes only small rotation during motion, the appearance estimation can also be good. In this case, the shape estimation is difficult due to the mostly homogeneous background similar to the object.

Fig.~\ref{fig:deblurring_shadow} is another interesting example of the deblatting behavior. The input frame is in the top left corner and the corresponding part from the high-speed camera is bellow. The object casts significant shadow. If we set the size of $F$ too small, the model cannot cope with the shadow and the estimated blur will contain artifacts in the locations of the shadow as is visible in the top row. If instead we make the support of $F$ sufficiently large, the estimated mask compensates for the shadow and the estimated blur is clean as shown in the bottom row.

\section{Conclusion}
\label{sec:conclusion}
We proposed a novel approach~--~Tracking by Deblatting~--~intended for sequences in which the object of interest undergoes non-negligible motion within a single frame, which needs to be specified by intra-frame trajectory rather than a single position.
The method is based on the observation that motion blur is directly related to the motion trajectory of the object. Blur is estimated by a complex method combining blind deblurring, image matting and shape estimation, 
followed by fitting a piecewise linear or quadratic curve that models physically plausible trajectories. As a result, we can precisely localize the object with higher temporal resolution than by conventional trackers. 

The proposed TbD tracker was evaluated on a newly created dataset of videos with ground truth obtained by a high-speed camera using a novel Trajectory-IoU metric that generalizes the traditional Intersection over Union and measures the accuracy of the intra-frame trajectory. The proposed method outperforms baseline techniques both in recall and trajectory accuracy. 

Due to the complexity of blind deblurring, the method is currently limited to objects that do not significantly change their perceived shape and appearance within a single frame, the method works best for approximately round and uniform objects.

{\small
\bibliographystyle{ieee}
\bibliography{main}
}

\end{document}